\documentclass[10pt, logo, copyright]{nv}

\usepackage{booktabs} 
\usepackage{caption}  
\usepackage{pifont}

\usepackage{amssymb}
\usepackage{subcaption}
\usepackage[table]{xcolor}
\usepackage{xcolor}   
\usepackage{multirow}
\usepackage{makecell} 
\usepackage{url}
\usepackage{booktabs}
\usepackage{graphicx}
\usepackage{animate}
\usepackage{xspace}
\usepackage{amsmath,amsfonts,bm}
\usepackage{enumitem}
\usepackage{cite}
\usepackage{wrapfig}
\usepackage{algorithm}
\usepackage{algorithmicx}
\usepackage{algcompatible}
\usepackage{algpseudocode}

\usepackage{cleveref}
\crefname{section}{Sec.}{Secs.}
\crefname{table}{Tab.}{Tabs.}
\crefname{figure}{Fig.}{Figs.}
\crefname{algorithm}{Alg.}{Algs.}

\definecolor{cvprblue}{rgb}{0.21,0.49,0.74}
\definecolor{iccvblue}{rgb}{0.21,0.49,0.74}
\definecolor{mitblue}{rgb}{0.88,0.95,0.96}
\definecolor{gold}{rgb}{0.75,0.6,0.12}
\colorlet{shadecolor}{gray!40}
\definecolor{mydarkred}{rgb}{0.8,0.02,0.02}

\newcolumntype{g}{>{\columncolor{mitblue}}c}
\newcolumntype{f}{>{\columncolor{mitblue}}l}
\newcolumntype{h}{>{\columncolor{mitblue}}r}
\newcolumntype{i}{>{\columncolor{gray}}c}

\newcommand{\eg}{\textit{e.g.,}\xspace}
\newcommand{\ie}{\textit{i.e.,}\xspace}
\newcommand{\myPara}[1]{\noindent\textbf{#1}}

\newcommand{\secref}[1]{Sec.~\ref{#1}}
\newcommand{\tabref}[1]{Table.~\ref{#1}}
\newcommand{\figref}[1]{Fig.~\ref{#1}}
\newcommand{\equref}[1]{Eq.~\ref{#1}}
\usepackage[most]{tcolorbox}

\title{DC-Gen: Post-Training Diffusion Acceleration with Deeply Compressed Latent Space} 

\author{
Wenkun He$^\dag$,
Yuchao Gu$^\dag$,
Junyu Chen$^\dag$,
Junyi Wu,
Wenhang Ge,
Dongyun Zou,
Yujun Lin,
Zhekai Zhang,
Haocheng Xi,
Muyang Li,
Ligeng Zhu,
Yuyang Zhao,
Jincheng Yu,
Junsong Chen,
Enze Xie,
Song Han,
Han Cai \\~\\
}

\vspace{-10pt}
\begin{abstract}
Existing latent diffusion models excel at visual generation and editing tasks, employing autoencoders to project RGB images and videos into latent spaces. However, their low-compression latents produce redundant and excessive visual tokens, limiting scalability to high-resolution content such as 4K images and 2160p videos.
One promising approach for acceleration is to leverage deeply compressed latent spaces produced by deep compression autoencoders. However, training industrial-scale models from scratch remains expensive and often fails to match the performance of pre-trained models. Alternatively, directly fine-tuning existing pre-trained models on deeply compressed latent spaces is appealing, but it introduces a critical challenge: a representation gap between the original latent space and the target compressed space, which can result in suboptimal convergence and even training instability.
DC-Gen offers the first solution to tackle this challenge. Specifically, we introduce a novel embedding alignment process to bridge the representation gap. Once the embedding spaces are aligned, only lightweight LoRA fine-tuning is required to unlock the pre-trained model’s full generative capability.
We validate the effectiveness of DC-Gen across tasks ranging from image and video generation to image editing. The resulting DC-Gen models achieve quality comparable to pre-trained models while delivering substantial speedups. For example, DC-Gen-FLUX reduces the latency of 4K image generation by 53.8$\times$ on a single NVIDIA H100 GPU.

\end{abstract}

\begin{document}

\maketitle
\begin{figure}[htbp]
    \centering
    \includegraphics[width=\linewidth]{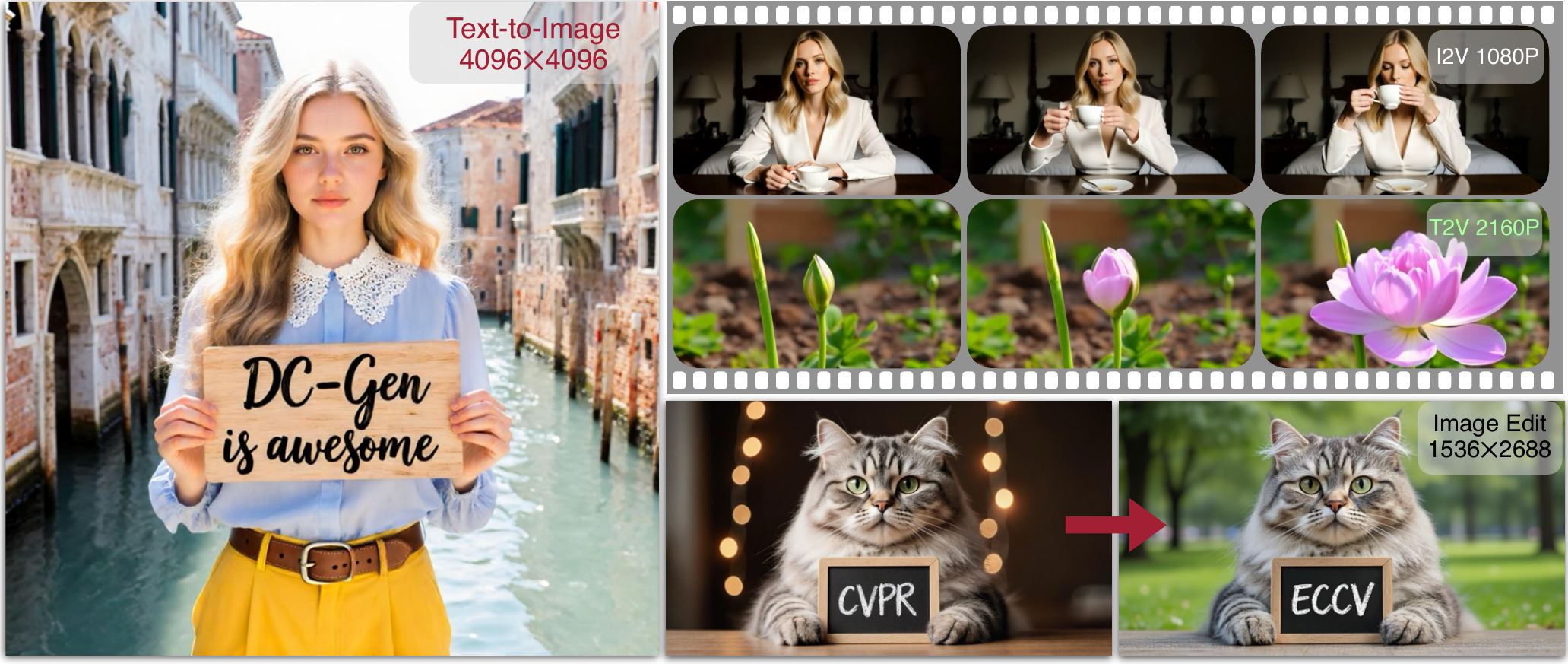}
    \vspace{-20pt}
    \caption{DC-Gen delivers high-quality, high-resolution visual generation across text-to-image, text-to-video, image-to-video, and image editing tasks, achieving up to 53.8$\times$ speedup over pre-trained diffusion models.
    }
    \vspace{-20pt}
    \label{fig:teaser}
\end{figure}




\section{Introduction}
\label{sec:intro}

Recent latent diffusion models~\cite{chen2024pixart, flux1kreadev2025, wan2025, wu2025qwenimagetechnicalreport} have achieved remarkable quality, but their slow inference speed remains a major bottleneck, particularly for high-resolution image and video generation. To address this issue, various acceleration techniques have been explored, including quantization~\cite{li2024svdquant}, few-step distillation~\cite{luo2023latent}, and feature caching~\cite{ma2024deepcache}. Nevertheless, a critical inefficiency remains: \textit{the inherent redundancy within the latent space}.
Most state-of-the-art image and video diffusion models~\cite{flux1kreadev2025, wan2025} use moderately compressed latent spaces. For instance, FLUX.1-Krea~\cite{flux1kreadev2025} employs an autoencoder with an 8$\times$ spatial compression ratio, representing a 4K image with 65,536 visual tokens. Similarly, Wan 2.1~\cite{wan2025} uses an autoencoder with $8\times$ spatial and $4\times$ temporal compression, resulting in 75,600 tokens for an 81-frame 720P video. These large token counts make inference computationally expensive, taking over three minutes to generate a single 4K image and approximately half an hour for an 81-frame 720P video on an NVIDIA H100 GPU.

To address redundancy in latent spaces, the DC-AE series~\cite{chen2024deep, chen2025dc} explores deeply compressed latent spaces. Specifically, DC-AE achieves reconstruction quality comparable to autoencoders with lower compression ratios (e.g., $8\times$) while employing spatial compression ratios of $32\times$ and $64\times$. Integrating such deeply compressed latents into existing diffusion models could yield substantial efficiency gains.
However, training high-quality latent diffusion models from scratch in deeply compressed latent spaces remains challenging due to substantial computational and data demands. Consequently, their potential for accelerating diffusion models remains largely underexplored.

To address this challenge, we introduce DC-Gen, an effective paradigm for adapting pre-trained diffusion models to deeply compressed latent spaces through cost-efficient post-training. Simply replacing the original latent space with a deeply compressed one creates a representation gap, which leads to suboptimal convergence or even training instability during end-to-end fine-tuning, as shown in \figref{fig:ablation_alignment_vis} and \figref{fig:ablation_alignment_metrics}. To overcome this issue, DC-Gen incorporates a lightweight \textbf{embedding alignment training} stage (\secref{sec:challenge_and_pipeline}), providing an ideal starting point for further training. Built on this aligned embedding space, only lightweight LoRA fine-tuning is required to recover the pre-trained model’s full capability. With DC-Gen, pre-trained diffusion models can be accelerated using deeply compressed latent spaces while preserving their original generation quality.

To validate the effectiveness and broad applicability of DC-Gen, we propose several additional techniques to apply it to a variety of diffusion models across image generation, video generation, and image editing tasks, including FLUX.1-Krea~\cite{flux1kreadev2025}, Z-Image-Turbo~\cite{imageteam2025zimageefficientimagegeneration}, Wan2.1~\cite{wan2025}, and Qwen-Image-Edit~\cite{wu2025qwenimagetechnicalreport}. The resulting DC-Gen models demonstrate substantial efficiency gains while maintaining quality comparable to their respective pre-trained models (\tabref{tab:mjhq_1024}, \tabref{tab:video_720}, \tabref{tab:image_edit_1024}). For example, DC-Gen-FLUX achieves approximately a 53.8$\times$ speedup in 4K image generation compared to the pre-trained FLUX.1-Krea, with the entire post-training process completed within 40 H100 GPU days (2.5 days on 16 GPUs). The contributions of DC-Gen are summarized as follows:
\begin{itemize}[leftmargin=*]
    \item We introduce DC-Gen, an effective acceleration paradigm for adapting pre-trained latent diffusion models to deeply compressed latent spaces. This approach leverages cost-efficient post-training to achieve substantial speedups without the overhead of training from scratch.
    \item We propose a novel embedding alignment training stage to bridge the representation gap when switching latent spaces, stabilizing training and enabling the pre-trained model to recover its full generation quality in the new compressed latent space.
    \item We validate the effectiveness of DC-Gen across a wide range of tasks, including image generation, video generation, and image editing. Our results demonstrate that DC-Gen preserves the performance of pre-trained models when adapting to deeply compressed latent spaces, while delivering substantial speedups for high-resolution generation.
\end{itemize}

\section{Related Work}
\subsection{Latent Diffusion Models}

Training diffusion models on the raw pixel space~\cite{saharia2022photorealistic} is computationally prohibitive. To address this, previous works explored training latent diffusion models~\cite{rombach2022high, peebles2023scalable, bao2023all, li2024playground, liu2024playground, blattmann2023align, blattmann2023stable, zhou2024upscale, xu2024easyanimate} on compressed latent spaces generated by autoencoders. State-of-the-art visual diffusion models~\cite{flux1kreadev2025, wan2025, wu2025qwenimagetechnicalreport, huang2024context,kumari2023multi, labs2025flux1kontextflowmatching,ruiz2023dreambooth} typically adopt an 8$\times$ compression ratio. However, this ratio results in redundant tokens for high-resolution image and video synthesis~\cite{zhang2025diffusion4k}. From the autoencoder perspective, later works have explored deeply compressed latent spaces~\cite{chen2024softvq, chen2024deep, chen2025masked, agarwal2025cosmos, wan2025, ma2025step, hacohen2024ltx, peng2025open}, improved latent spaces with regularizations~\cite{yu2024representation, yao2025vavae, gu2024rethinking, chen2025dc, yao2025reconstruction, kouzelis2025eq, skorokhodov2025improving, gu2025long}, or simply increasing latent channel numbers~\cite{dai2023emu,flux2024, esser2024scaling}. Some other works~\cite{blattmann2022retrieval, chen2022re, ye2023ip, suvorov2022resolution, yang2023paint} introduce additional modules or computation processes to enhance the quality of latent diffusion. DC-Gen neither trains a new autoencoder, nor introduces any additional procedures. Instead, leveraging existing pre-trained models, it achieves acceleration through efficient and stable post-training with deeply compressed latent spaces.

\subsection{Diffusion Model Acceleration}
A well-known drawback of diffusion models is their inefficient multi-step inference process, which has motivated numerous works to improve their inference efficiency. One series of works focuses on training-free acceleration, including the design of improved ODE solvers~\cite{songdenoising,lu2022dpm,lu2022dpm2,zheng2023dpm,zhangfast,zhanggddim,zhao2024unipc}, model quantization~\cite{li2023q,li2024svdquant,li2025radial,fang2024structural,zhang2024sageattention,zhang2024sageattention2,tian2024qvd,zhao2024vidit,chen2025q,huang2025qvgen,zhao2025paroattention,li2025dvd} and sparsity~\cite{xi2025sparse,yang2025sparse,zhang2025fast,xia2025training,tan2025dsv,li2025radial,zhang2025spargeattn,zhao2025paroattention}, and efficient inference systems~\cite{ma2024deepcache,shih2024parallel,tang2024accelerating,li2024distrifusion,wang2024pipefusion,wu2025quantcache}. 
Another series of works requires post-training for acceleration, like few-step distillation~\cite{luo2023latent,yin2024improved,yin2024one,wang2023videolcm,lin2024animatediff,li2024t2v,zhai2024motion,yin2024slow,wang2024animatelcm,zhang2024sf,mao2025osv,lin2025diffusion} and guidance distillation~\cite{flux2024,meng2023distillation,salimans2022progressive}.
In addition to accelerating pre-trained models, some works explore efficient architectures \cite{li2024snapfusion,liu2024linfusion,cai2024condition,ma2024sit,xie2024sana,xie2025sana,cai2023efficientvit} or better diffusion trajectory modeling \cite{frans2024one,zhou2025inductive,geng2025mean}. DC-Gen falls into the category of post-training acceleration \cite{gu2025jet}, which explores a new perspective by accelerating pre-trained diffusion models using deeply compressed latent spaces.

While effective deeply compressed latent spaces~\cite{chen2024deep,chen2025dc} have been developed, training high-quality visual diffusion models from scratch in these latent spaces remains expensive. Therefore, DC-Gen aims to accelerate pre-trained diffusion models in deeply compressed latent spaces through a cost-efficient post-training process rather than costly pre-training.

\section{Method}
\subsection{Motivation and Overview}
The core objective of DC-Gen is to accelerate pre-trained latent diffusion models by migrating from the original 8$\times$ spatial compression latent space to a deeply compressed alternative (32$\times$ or 64$\times$). By leveraging a post-training approach, DC-Gen circumvents the prohibitive costs and the high-quality data availability challenges typically associated with training high-fidelity models from scratch.

The post-trained DC-Gen models offer several key advantages. (1) They use fewer visual tokens to represent images or videos, which leads to significantly faster inference speeds. (2) They inherit the pre-trained model's generation capability even without requiring the original pre-training dataset. (3) Due to the substantial token reduction, they enable considerably more efficient fine-tuning for downstream applications, such as RL training or LoRA fine-tuning, compared to the pre-trained models.

Despite these notable merits, adapting a pre-trained diffusion model to a new deeply compressed latent space remains non-trivial. Previous works~\cite{peng2025open} have reported quality degradation under na\"{\i}ve adaptation (\secref{sec:discussion}). To address this issue, DC-Gen introduces a novel embedding alignment stage prior to end-to-end fine-tuning. We first present the preliminaries in \secref{sec:preliminary}, followed by the embedding alignment training in \secref{sec:challenge_and_pipeline}. We then conduct extensive ablation studies to validate the necessity of embedding alignment across a broad range of tasks and diffusion models in \secref{sec:discussion}. Finally, we present specific techniques and implementation details for applying DC-Gen to different visual generation tasks in \secref{sec:application}. An overview of the DC-Gen framework is shown in Fig.~\ref{fig:pipeline}.

\subsection{Preliminary}
\label{sec:preliminary}
We first introduce the key concepts of latent image diffusion models.
\begin{itemize}[leftmargin=*]
\item \textbf{Compression Ratios and Channel Numbers.} The \textbf{spatial compression ratio} $f$ is defined as the ratio of the original pixel dimensions $H \times W$ to the latent space dimensions $\frac H f\times \frac W f$, while the \textbf{channel number} $c$ denotes the latent dimension. For example, a standard $f8c16$ image autoencoder compresses an image of size $3\times H\times W$ into a latent of size $c\times \frac H f\times \frac W f=16\times \frac{H}{8}\times \frac{W}{8}$.

\item \textbf{Converting Latents into Visual Tokens.} The transformer backbone of a latent diffusion model operates on a sequence of visual tokens. To embed the latents into such tokens, a patch-based tokenization process is used. The \textbf{patch size} $p$ divides the latents into non-overlapping patches of size $p \times p$, each of which is embedded into a single token. The conversion between latents and tokens is performed by two lightweight modules in the latent diffusion model: The \textbf{patch embedder} that transforms the latents into embedding tokens, and the \textbf{output layer} that maps the embeddings back to the latent space (\figref{fig:pipeline}(a)). These modules are tied to the channel number and the patch size, which cannot be reused when adapting to latent spaces with different shapes.

\item \textbf{Token Counts.} The total number of visual tokens for a diffusion model can be calculated using both compression factors $f$ and $p$. The formula is as follows:
\begin{equation}
\text{Token Counts} = \left(\frac{H}{f \cdot p}\right) \times \left(\frac{W}{f \cdot p}\right),    \nonumber
\end{equation}
where $H$ and $W$ are the original pixel dimensions. For instance, an $f8p2$ latent diffusion model converts a $1024 \times 1024$ image into $4096$ tokens.
\end{itemize}

For video generation, the additional temporal dimension is further reduced by the \textbf{temporal compression ratio} $t$, which converts $T$ video frames into $\frac T t$ latent frames. For instance, an $f8t4c16$ video autoencoder compresses a video of size $3\times T\times H\times W$ into a latent of size $c\times \frac T t\times \frac H f\times \frac W f=16\times \frac T 4 \times \frac H 8\times \frac W 8$.

\subsection{Challenge, Analysis, and the DC-Gen Pipeline}
\label{sec:challenge_and_pipeline}
To replace the latent space with a deeply compressed one while avoiding the prohibitive costs of training from scratch, a straightforward approach is to reuse the pre-trained transformer backbone weights and randomly initialize the patch embedder and output layer which are responsible for adapting the model to the new latent space, followed by end-to-end fine-tuning. We refer to this baseline approach as \textbf{direct fine-tuning}. 
In practice, however, it suffers from severe performance degradation. Even as training appears to converge, visual quality remains significantly inferior to the pre-trained model. In extreme cases, the training process even becomes unstable, yielding outputs that are virtually indistinguishable from noise (\eg, the video generation results in \figref{fig:ablation_alignment_vis}).

\begin{figure*}[t]
    \centering
    \vspace{-.1in}
    \includegraphics[width=0.9\linewidth]{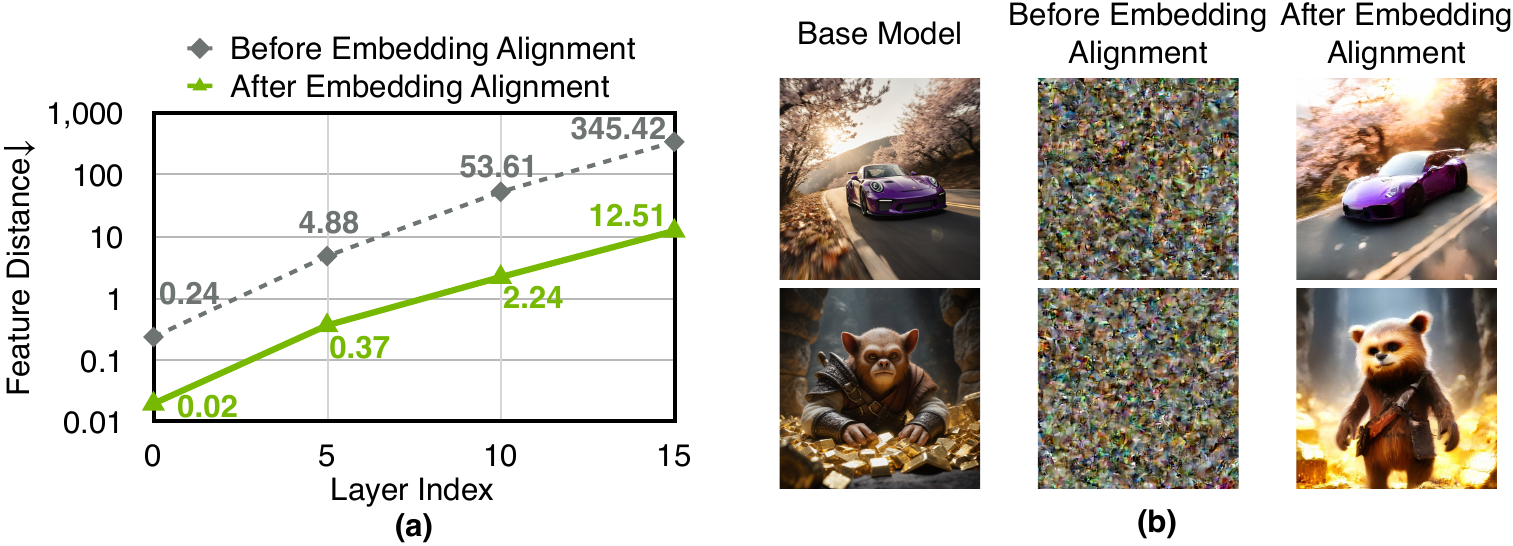}
    \vspace{-.1in}
    \caption{
   (a) Embedding alignment significantly reduces the per-layer feature distance between the pre-trained diffusion model.
   (b) After embedding alignment, the model can generate semantically correct images in the new latent space, even without fine-tuning the diffusion backbone. Experiments are conducted on FLUX.1-Krea~\cite{flux1kreadev2025}.}
    \label{fig:analysis}
    \vspace{-.1in}
\end{figure*}

We attribute this failure primarily to the \textbf{representation gap} between the original and new latent spaces. As shown in \figref{fig:analysis}(a), we measured the per-layer feature distance between the embedding tokens generated by the pre-trained diffusion model and those from the newly initialized one. The deeply compressed embedding tokens produced by the random patch embedder deviate from the pre-trained representation, and this discrepancy grows exponentially through the layers of the transformer backbone. This divergence prevents the transformer backbone from effectively recovering its inherent pre-trained capabilities, leading to the observed performance degradation.

\begin{figure*}[t]
    \centering
    \includegraphics[width=\linewidth]{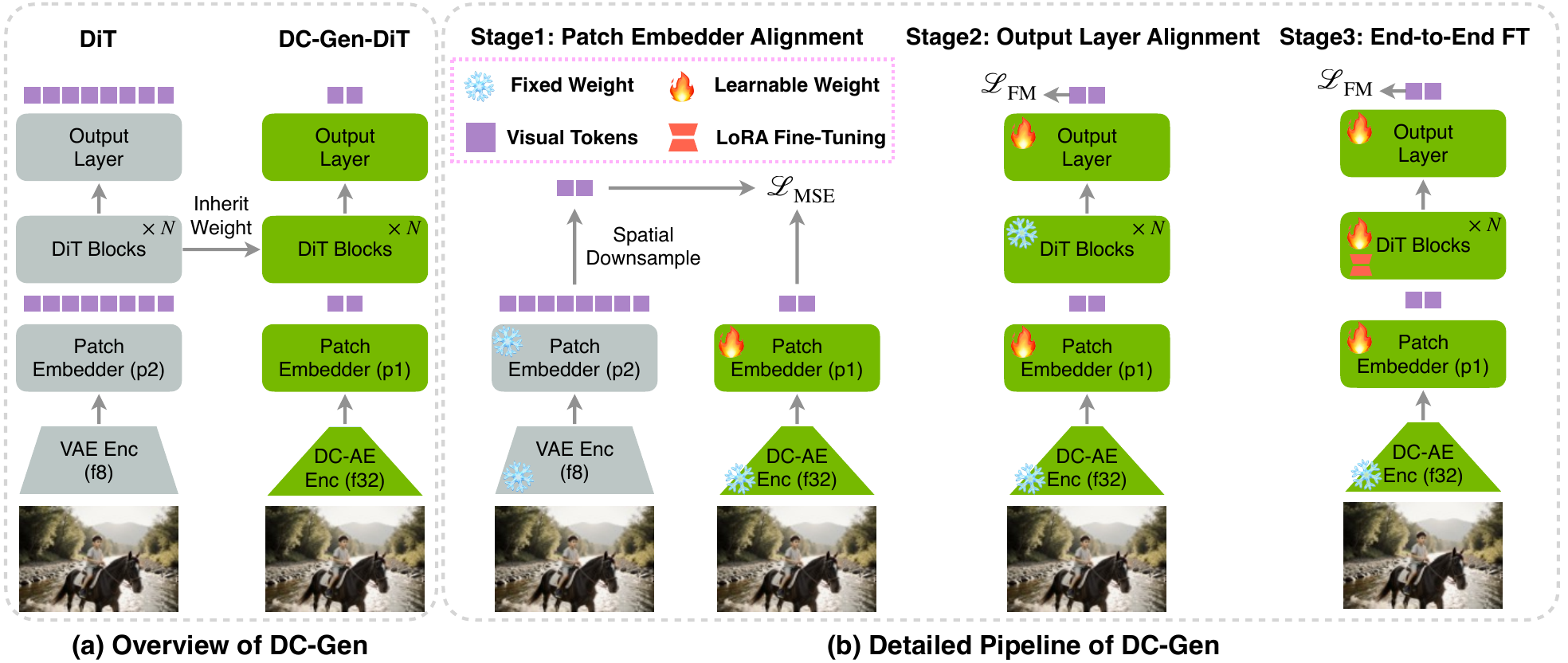}
    \caption{
   \textbf{Illustration of DC-Gen.} 
   (a) DC-Gen accelerates diffusion models by adapting pre-trained diffusion models to deeply compressed latent spaces (\eg, DC-AE~\cite{chen2024deep}).
   (b) DC-Gen introduces a patch embedder alignment stage to bridge the representation gap between the original latent space and the target deeply compressed space, followed by a lightweight output-layer alignment stage. This design accelerates convergence, mitigates training instability, and preserves the capability of the pre-trained model.
   }
   \vspace{-.2in}
\label{fig:pipeline}
\end{figure*}

To bridge this gap, we propose a lightweight embedding alignment stage. This process aligns the patch embedder and output layer with the pre-trained model’s embedding space prior to end-to-end fine-tuning, as depicted in \figref{fig:pipeline}(b).

\vspace{0.1cm}
\noindent\textbf{Patch Embedder Alignment.} 
Let $\mathbf{x}$ be an input RGB image (or video). We denote its latent representations in the original and deeply compressed latent spaces as $\mathbf z_o=E_o(\mathbf{x})$ and $\mathbf z_d=E_d(\mathbf{x})$, respectively.

For the pre-trained patch embedder $\mathbf{e}_o$ corresponding to the original latent space, we denote the encoded embedding tokens as $\mathbf{e}_o(\mathbf z_o) \in \mathbb{R}^{H_o \times W_o \times D}$ (or $\mathbf{e}_o(\mathbf z_o) \in \mathbb{R}^{T_o \times H_o \times W_o \times D}$ for videos), where $H_o=\frac{H}{f_o p_o}$ and $W_o=\frac{W}{f_o p_o}$. Here, $f_o$ denotes the spatial compression ratio of the original latent space, $p_o$ denotes the patch size used in the pre-trained model, and $D$ is the hidden dimension of the diffusion transformer backbone.

Simliarly, for the randomly initialized patch embedder $\mathbf{e}_{d,\phi}$ corresponding to the deeply compressed latent space, we denote the encoded embedding tokens as $\mathbf{e}_{d,\phi}(\mathbf z_d) \in \mathbb{R}^{H_d \times W_d \times D}$ (or $\mathbf{e}_{d,\phi}(\mathbf z_d) \in \mathbb{R}^{T_d \times H_d \times W_d \times D}$ for videos). Here $H_d=\frac{H}{f_d p_d}$ and $W_d=\frac{W}{f_d p_d}$, with $f_d$ denoting the spatial compression ratio of the deeply compressed latent space and $p_d$ denoting the new patch size. Note that $f_d p_d > f_o p_o$, resulting in a more compact representation.

We spatially downsample the pre-trained embedding $\mathbf{e}_o(\mathbf z_o)$ to match the dimensions of $\mathbf{e}_d(\mathbf z_d)$, denoted as $\hat{\mathbf{e}}_o(\mathbf z_o)$. Then, we optimize the parameters $\phi$ to minimize the distance between the two embeddings:
\begin{equation}
    \mathcal{L}_{\mathrm{MSE}}(\phi) = \mathbb E_{\mathbf x}\Vert \mathbf{e}_{d,\phi}(\mathbf z_d) - \hat{\mathbf{e}}_o(\mathbf z_o) \Vert_2^2 = \mathbb E_{\mathbf x}\Vert \mathbf{e}_{d,\phi}(E_d(\mathbf x)) - \hat{\mathbf{e}}_o(E_o(\mathbf x)) \Vert_2^2.
\label{eq:embedder_align}
\end{equation}

\vspace{0.1cm}
\noindent\textbf{Output Layer Alignment.} After training the patch embedder, the output layer remains randomly initialized. To further align the output representations, we jointly fine-tune both the patch embedder $\mathbf e_\phi$ and the output layer $\mathbf h_\theta$ for a few steps while keeping the transformer backbone frozen. The loss is the standard flow-matching objective~\cite{lipman2022flow,liu2022flow}:
\begin{equation}
\label{eq:fm}
\mathcal{L}_{\mathrm{FM}}(\phi,\theta) = \mathbb{E}_{\mathbf{z}_0, \boldsymbol{\epsilon}, t} \left[ \| \mathbf{v}_{\phi, \theta}(\mathbf{z}_t, \mathbf{c}, t) - \mathbf{v}_t \|_2^2 \right],
\end{equation}
where $\mathbf{z}_t = (1-t)\mathbf{z}_0 + t\boldsymbol{\epsilon}$ is calculated based on the clean latent sample $\mathbf{z}_0$, the noise sample $\boldsymbol{\epsilon}\sim\mathcal N(0, I)$, and the timestep $t\in[0, 1]$. $\mathbf c$ denotes the text condition, and the ground truth velocity is given by $\mathbf{v}_t = \boldsymbol{\epsilon} - \mathbf{z}_0$.

As shown in \figref{fig:analysis}(a), after the embedding alignment of these two lightweight modules, the per-layer feature distance is substantially reduced. As shown in \figref{fig:analysis}(b), the model can already generate semantically reasonable samples in the new latent space without fine-tuning the diffusion transformer backbone.

\vspace{0.1cm}
\noindent\textbf{End-to-End Fine-Tuning.}
With embedding alignment, the pre-trained diffusion backbone is effectively synchronized with the new latent space via the optimized patch embedder and output layer. We then proceed with end-to-end fine-tuning of the entire diffusion model. Since the alignment stage largely preserves the foundation capabilities of the pre-trained model, only a brief period of LoRA fine-tuning is required to further refine details and restore the original generation quality. As detailed in \secref{sec:appendix_training_cost}, DC-Gen's three-stage post-training pipeline significantly reduces the computational cost compared to training state-of-the-art models from scratch.

\subsection{Discussion and Ablation Study on Embedding Alignment}
\label{sec:discussion}
To the best of our knowledge, DC-Gen is \textit{the first post-training pipeline to effectively enable adaptation to deeply compressed latent spaces}. Prior to DC-Gen, some image diffusion models switch latent spaces to alternative ones with identical spatial compression ratios and channel numbers for improved reconstruction quality. For example, PixArt-$\Sigma$~\cite{chen2024pixart} adapts SD-VAE-f8c4 to SDXL-VAE-f8c4. In such cases, the pre-trained patch embedder and output layer can be directly loaded due to the matching configurations.
In contrast, DC-Gen focuses on adaptation across latent spaces with different configurations. OpenSora-2.0~\cite{peng2025open} proposes a more analogous scenario, adapting from an f8 to an f32 latent space. However, their approach is functionally equivalent to \textbf{direct fine-tuning}. Consequently, they report that ``video outputs appear blurry'' (\cite{peng2025open}, Sec. 4.3) and that their ``fast video generation model underperforms the original'' (\cite{peng2025open}, Sec. 4.3), aligning with our observations.

\begin{figure}[htbp]
    \centering
    \vspace{-.3in}
    \includegraphics[width=\linewidth]{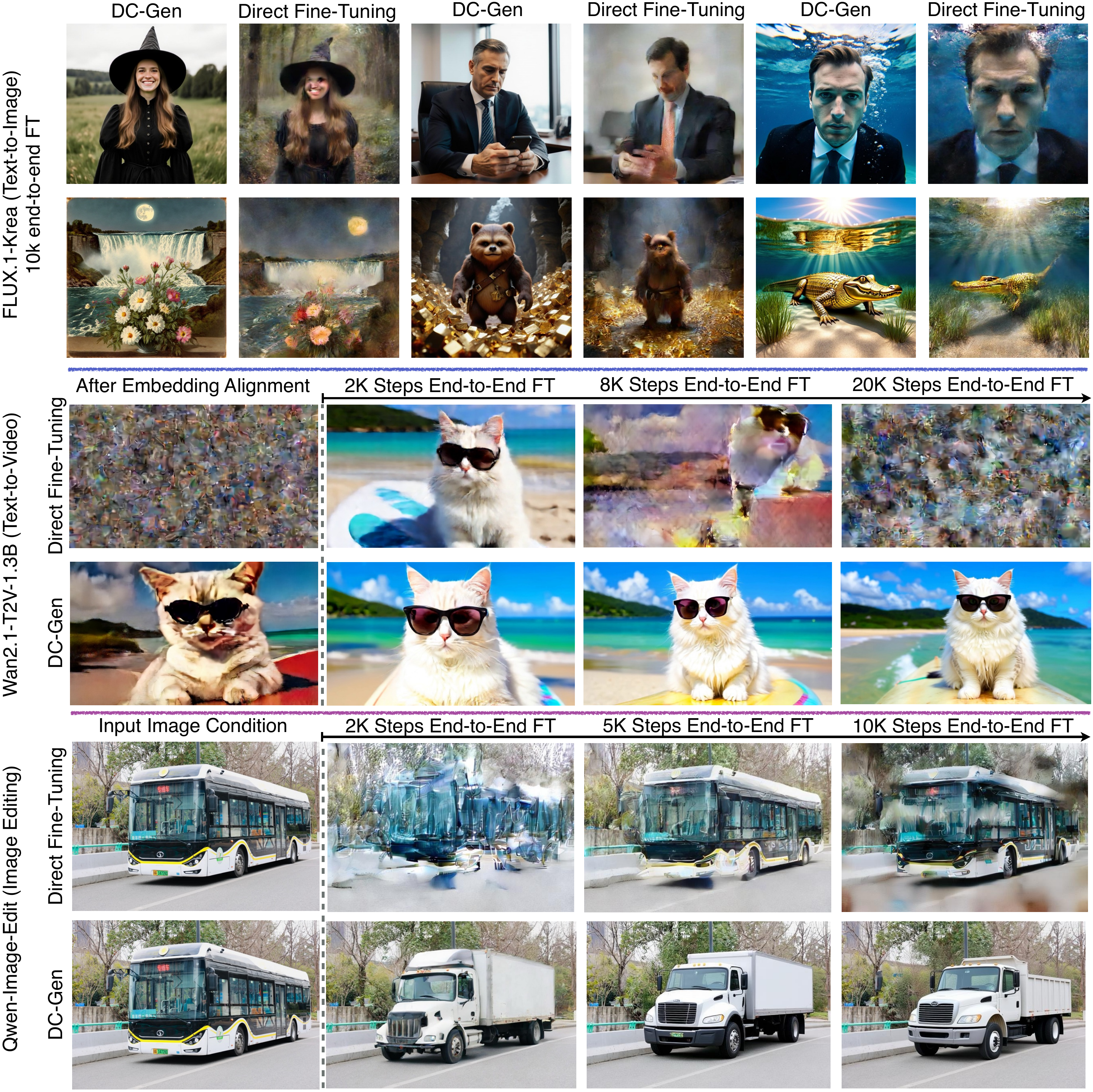}
    \vspace{-.25in}
    \caption{
         \textbf{Visual Comparison of Latent Space Adaptation Methods.} }
    \label{fig:ablation_alignment_vis}
    \vspace{-.25in}
\end{figure}

\begin{figure}[htbp]
    \centering
    \includegraphics[width=\linewidth]{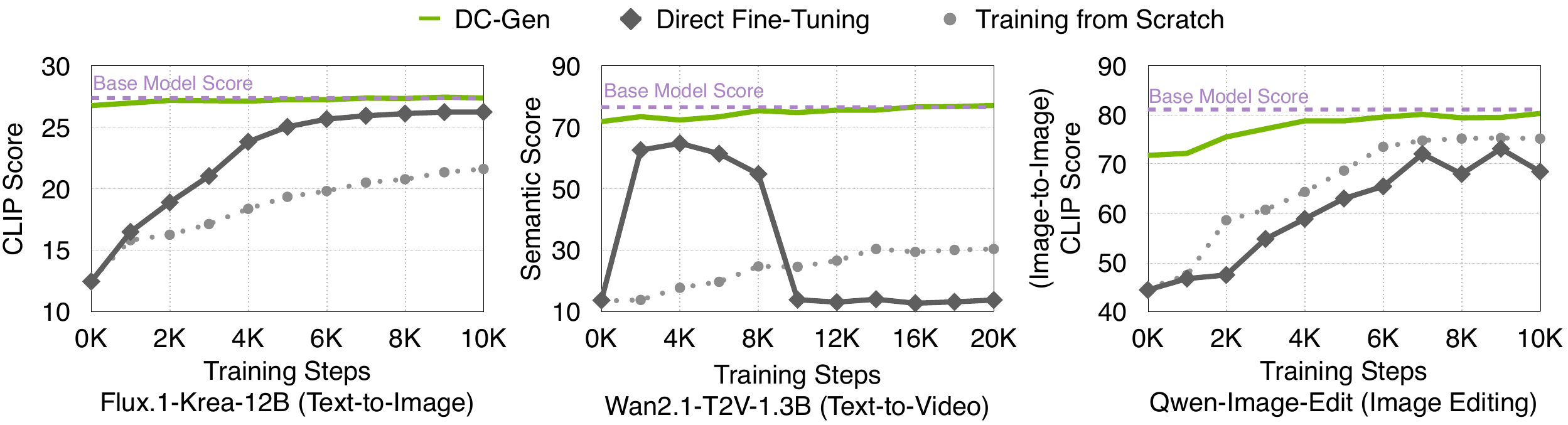}
    \vspace{-.25in}
    \caption{
         \textbf{Training Curve Comparison of Latent Space Adaptation Methods.}}
    \label{fig:ablation_alignment_metrics}
    \vspace{-.15in}
\end{figure}

The success of DC-Gen lies in its innovative embedding alignment, which provides a stable starting point for subsequent end-to-end fine-tuning while preserving the capability in the pre-trained backbone. We conduct an ablation study on the embedding alignment, comparing DC-Gen with \textbf{direct fine-tuning} and \textbf{training from scratch} (Randomly initialize the whole backbone and conduct end-to-end fine-tuning), as shown in \figref{fig:ablation_alignment_vis} and \figref{fig:ablation_alignment_metrics}. We use DC-AE-f32 to provide our target compressed latent space.

Without embedding alignment, \textbf{direct fine-tuning} degrades the visual quality of different models to varying degrees. For FLUX.1-Krea~\cite{flux1kreadev2025}, the end-to-end fine-tuning process remains relatively stable, yet the resulting image details are noticeably blurrier than those produced by the full DC-Gen pipeline. For Qwen-Image-Edit~\cite{wu2025qwenimagetechnicalreport}, \textbf{direct fine-tuning} manages to produce distinguishable structures, but details are severely distorted. Furthermore, the model tends to directly replicate the input image, failing to accomplish effective editing. In the most extreme case, Wan2.1-T2V-1.3B~\cite{wan2025} suffers from training instability: the training collapses after some steps, and the generated videos become nearly noise. 

In contrast, benefiting from a well-aligned initialization, DC-Gen requires only a few stable end-to-end fine-tuning steps to achieve comparable visual quality to the pre-trained model. Detailed comparisons are provided in \secref{sec:exp}.

In \secref{sec:appendix_embedding_alignment}, we present additional experiments to separate the effects of patch embedder alignment and output layer alignment, showing that the former is crucial for recovering the pre-trained model’s capabilities, while the latter speeds up convergence during end-to-end fine-tuning. We also compare DC-Gen with other training stabilization techniques, demonstrating that DC-Gen is more effective for adapting diffusion models to compressed latent spaces.

\subsection{Additional Techniques and Implementation Details}
\label{sec:application}
\vspace{0.1cm}
\noindent\textbf{DC-Gen for Text-to-Image Generation.}
We employ FLUX.1-Krea-12B~\cite{flux1kreadev2025} and Z-Image-Turbo~\cite{imageteam2025zimageefficientimagegeneration} as our pre-trained text-to-image models. 

To accelerate inference, they utilize guidance-distillation, which distills the Classifier-Free Guidance (CFG) scale~\cite{ho2022classifier} as a condition embedding. Adopting the standard flow-matching objective (\equref{eq:fm}) for training will result in a biased velocity estimation. To address this, we derive a corrected training objective:
\begin{equation}
\mathcal{L}_{\text{FM}}^{\text{guide}}
=
\mathbb{E}_{\mathbf{z}_0, \boldsymbol{\epsilon}, t, w}
\left[
\left\|
\frac{1}{1+w}
\left(
\mathbf{v}_{\eta}(\mathbf{z}_t, \mathbf c, t, w)
+
w\mathbf{v}_{\eta}(\mathbf{z}_t, \hat{\mathbf c}, t, w)
\right)
-
\mathbf{v}_t
\right\|_2^2
\right],
\end{equation}
where $\mathbf{v}_{\eta}$ is the distilled model, $\mathbf c$ is the text condition, $\hat{\mathbf c}$ is the negative text condition, $w$ is the guidance scale, and $\mathbf{v}_t$ is the ground truth velocity. We provide detailed derivation and ablation study in \secref{sec:guidance_distilled}.

\vspace{0.1cm}
\noindent\textbf{DC-Gen for Video Generation.}
To adapt existing video diffusion models to deeply compressed latent spaces, we develop DC-AE-V based on DC-AE~\cite{chen2024deep}, with spatial compression ratio $f=32$, temporal compression ratio $t=4$, and channel number $c=32$. 
The core design is a novel \textbf{chunk-causal} temporal modeling strategy, which integrates bidirectional information flow within chunks, and causal information flow across chunks.
This design substantially improves reconstruction quality under deep compression settings while preserving generalization to longer videos during inference. Details are analyzed in \secref{sec:appendix_dc_ae_v}.

Compared to T2I generation, the 3-stage DC-Gen pipeline for videos operates on the 3D video tensors, while the rest remains unchanged. For I2V, the Wan2.1 model incorporates the image condition by concatenating it with the latent. Since Wan2.1-VAE and DC-AE-V employ different temporal modeling designs (causal v.s. chunk-causal), we replicate the input image four times and append blank frames to match the video shape. We then encode these frames using DC-AE-V and concatenate the resulting image condition with the latent, which can subsequently be processed in the same manner as in T2V models.

\section{Experiments}
\label{sec:exp}
We validate the effectiveness of DC-Gen across multiple visual generation tasks, including text-to-image generation (\secref{T2I Gen}), text-to-video and image-to-video generation (\secref{Video gen}), and image editing (\secref{Image Editing}). 

We benchmark the inference latency of all models using TensorRT\footnote{\url{https://github.com/NVIDIA/TensorRT}} on a single H100 GPU. For simplicity, we focus exclusively on the transformer backbone, as it constitutes the primary efficiency bottleneck.

\subsection{Text-to-Image Generation}
\label{T2I Gen}
\noindent\textbf{Implementation Details.}
We train all models using 16 NVIDIA H100 GPUs. FLUX.1-Krea-12B~\cite{flux1kreadev2025} and Z-Image-Turbo~\cite{imageteam2025zimageefficientimagegeneration} serve as the pre-trained backbones. We replace their f8 autoencoders with DC-AE-f32 for 1K-resolution image generation. To further improve efficiency at 2K and 4K resolutions, we apply DC-Gen to adapt from DC-AE-f32 to DC-AE-f64. In the end-to-end fine-tuning stage, we apply LoRA~\cite{hu2022lora} with a rank and alpha of 256 to all linear modules. The detailed training hyperparameters are provided in \secref{sec:hyper_params}.

\vspace{0.1cm}
\noindent\textbf{Dataset and Evaluation Protocol.}
For both models, the post-training dataset comprises real-world images from Pexels\footnote{\url{https://www.pexels.com}} and Unsplash~\cite{unsplash_dataset}, and synthetic data generated by the pre-trained model.
For quantitative evaluation at $1024\times 1024$ resolution, we use the MJHQ-30K~\cite{li2024playground} dataset and report CLIP-Score and GenEval metrics, following common practice~\cite{xie2024sana}. For 4K evaluation, we use Aesthetic-4K~\cite{zhang2025diffusion4k}, the current state-of-the-art benchmark for 4K image generation.

\begin{figure}[h]
    \centering
    \vspace{-10pt}
    \includegraphics[width=\linewidth]{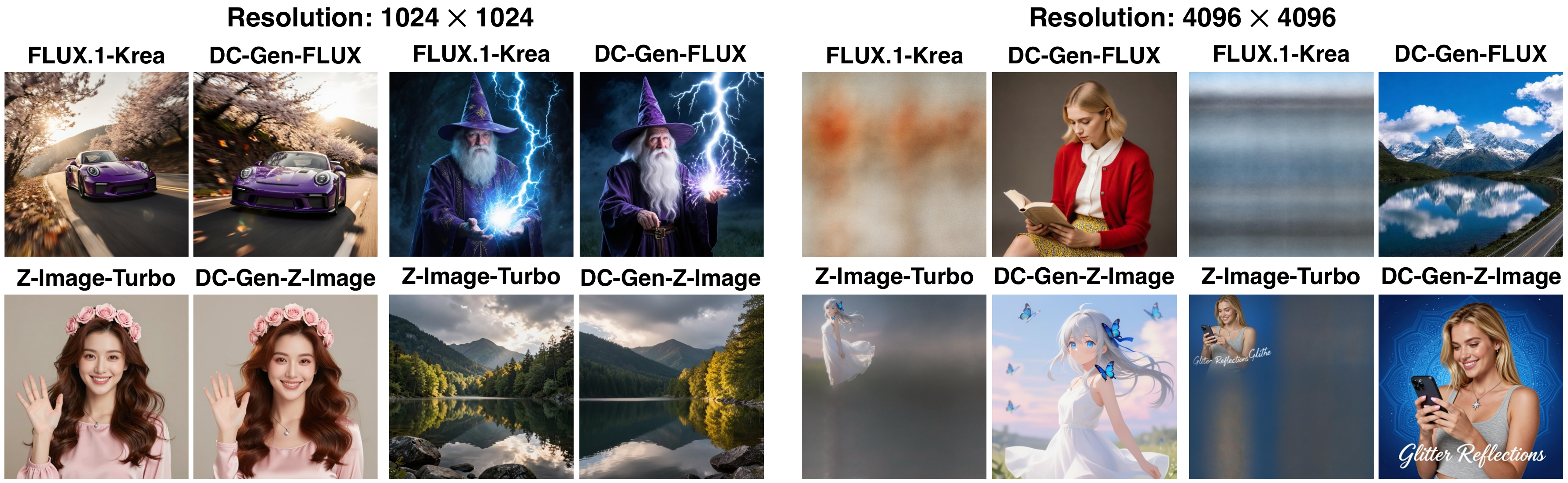}
    \vspace{-20pt}
    \caption{
   \textbf{Visual Results on Text-to-Image Generation.} }
    \label{fig:t2i_samples_main_text}
\end{figure}

\vspace{0.1cm}
\noindent\textbf{Experimental Results.}
We demonstrate qualitative comparisons between DC-Gen-FLUX, DC-Gen-Z-Image, and their pre-trained models at resolutions of 1K and 4K in \figref{fig:t2i_samples_main_text} and \secref{sec:appendix_t2i_samples}. DC-Gen shows similar quality and text alignment to the pre-trained model, successfully preserving pre-trained capabilities when shifting to a deeply compressed latent space. Note that FLUX.1-Krea-12B~\cite{flux1kreadev2025} and Z-Image-Turbo~\cite{imageteam2025zimageefficientimagegeneration}, the pre-trained models, do not natively support 4K image generation, likely due to high training costs that prevent the pre-training. The quantitative results are summarized in \tabref{tab:mjhq_1024} and \tabref{tab:t2i_4K}. DC-Gen-FLUX and DC-Gen-Z-Image achieve comparable quantitative results to corresponding pre-trained models, but with an approximately 4$\times$ latency reduction at 1K resolution, over 50$\times$ latency reduction at 4K resolution.

\begin{table*}[h!]
\vspace{-5pt}
\caption{\textbf{Text-to-Image Generation Results.} }
\small\centering\setlength{\tabcolsep}{3pt}
\vspace{-10pt}
\resizebox{0.75\linewidth}{!}{
\begin{tabular}{g | g | g | g  g }
\toprule
\multicolumn{5}{l}{\textbf{Text-to-Image Generation Results on MJHQ-30K 1024$\times$1024}} \\
\midrule
\rowcolor{white} & & \textbf{Latency} & & \\
\rowcolor{white} \multirow{-2}{*}{\textbf{Models}} & \multirow{-2}{*}{\textbf{Autoencoder}} & \textbf{(s)} $\downarrow$ & \multirow{-2}{*}{\textbf{CLIP Score} $\uparrow$} & \multirow{-2}{*}{\textbf{GenEval} $\uparrow$} \\
\midrule
\rowcolor{white} LUMINA-Next~\cite{gao2024lumina-next} & - & 9.13 & 26.84 & 0.46 \\
\rowcolor{white} SD3-medium~\cite{esser2024scaling} & - & 4.37 & 27.83 & 0.62 \\
\rowcolor{white} Hunyuan-DiT~\cite{li2024hunyuandit} & - & 18.16 & 28.19 & 0.63 \\
\rowcolor{white} PixArt-$\Sigma$~\cite{chen2024pixart} & - & 2.72 & 28.26 & 0.54 \\
\rowcolor{white} SDXL~\cite{podell2023sdxl} & - & 6.48 & 29.03 & 0.55 \\
\rowcolor{white} PlayGround~\cite{li2024playground} & - & 5.30 & 29.13 & 0.56 \\
\midrule
\midrule
\rowcolor{white} FLUX.1-Krea-12B~\cite{flux1kreadev2025} & FLUX-VAE-f8c16 & 4.06 & 27.93 & 0.69 \\
DC-Gen-FLUX & DC-AE-f32c32~\cite{chen2024deep} & 0.88 & 28.02 & 0.71 \\
\midrule
\midrule
\rowcolor{white} Z-Image-Turbo~\cite{imageteam2025zimageefficientimagegeneration} & Z-Image-VAE-f8c16 & 1.92 & 27.61 & 0.84 \\
DC-Gen-Z-Image & DC-AE-f32c32~\cite{chen2024deep} & 0.47 & 27.58 & 0.84 \\
\bottomrule
\end{tabular}
}
\label{tab:mjhq_1024}
\end{table*}
\begin{table}[t]
\vspace{-5pt}
\caption{\textbf{4K-Resolution Image Generation Results on Diffusion-4K \cite{zhang2025diffusion4k}.}}
\vspace{-10pt}
\footnotesize\centering\setlength{\tabcolsep}{3pt}
\resizebox{0.8\linewidth}{!}{
\begin{tabular}{g | g | g | g g}
\toprule
\rowcolor{white} & & \textbf{Latency} &  \\
\rowcolor{white} \multirow{-2}{*}{\textbf{Models}} & \multirow{-2}{*}{\textbf{Steps}} & \textbf{(s) $\downarrow$} & \multirow{-2}{*}{\textbf{CLIP Score} $\uparrow$}  & \multirow{-2}{*}{\textbf{Aesthetic Score} $\uparrow$}  \\\midrule
\rowcolor{lightgray} FLUX.1-Krea-12B~\cite{flux1kreadev2025} & 20 & 218.81 & - & - \\
\rowcolor{lightgray} Z-Image-Turbo~\cite{imageteam2025zimageefficientimagegeneration} & 9 & 159.23 & - & - \\
\midrule
\rowcolor{white} & 20 & 23.76 & 32.79 & 5.94 \\
\rowcolor{white} \multirow{-2}{*}{Diffusion-4K (FLUX.1-WLF) \cite{zhang2025diffusion4k}} & 50 & 59.42 & 33.12 & 6.06 \\
\midrule
DC-Gen-FLUX & 20 & 4.04 & \textbf{35.07} & \textbf{6.31} \\
DC-Gen-Z-Image & 9 & \textbf{1.91} & 34.23 & 6.11 \\
\bottomrule
\end{tabular}
}
\label{tab:t2i_4K}
\end{table}

To measure from a perceptual perspective, we also conducted a human evaluation. To ensure unbiased assessment, we use Amazon Mechanical Turk, a crowdsourcing platform. Our evaluation covers three common T2I use cases: general, person, and anime. Each domain includes 25 test prompts, resulting in 75 total prompts. For each prompt, we generate an image pair using DC-Gen and the pre-trained model, and present the pair to 20 users. Sample images and the user interface example are provided in \secref{sec:human_study_t2i}. Users answer two questions: which image has higher visual quality, and which image aligns more closely with the prompt. \tabref{tab:user_study} summarizes the results. DC-Gen achieves comparable visual quality and text alignment to the pre-trained model across all domains.

\begin{table}[h!]
\vspace{-5pt}
\caption{\textbf{Human Evaluation Results.} We conducted a human evaluation on Amazon Mechanical Turk, where DC-Gen achieved performance on par with the pre-trained models in both visual quality and text alignment.}
\vspace{-5pt}
\small\centering\setlength{\tabcolsep}{2.5pt}
\resizebox{0.85\linewidth}{!}{
\begin{tabular}{l | c c c | c c c}
\toprule
& \multicolumn{3}{c|}{\textbf{Visual Quality}} & \multicolumn{3}{c}{\textbf{Text Alignment}} \\\cmidrule{2-7}
& \textbf{DC-Gen-FLUX} & & \textbf{FLUX.1-Krea} & \textbf{DC-Gen-FLUX} & & \textbf{FLUX.1-Krea} \\
\multirow{-3}{*}{\textbf{Domain}} & \textbf{is better} & \multirow{-2}{*}{\textbf{Equal}} & \textbf{is better}  & \textbf{is better} & \multirow{-2}{*}{\textbf{Equal}} & \textbf{is better} \\
\midrule
General & 47.4\% & 9.6\% & 43.0\% & 46.4\% & 13.0\% & 40.6\% \\
Person & 46.8\% & 8.6\% & 44.6\% & 43.4\% & 13.6\% & 43.0\% \\
Anime & 48.0\% & 7.4\% & 44.6\% & 47.4\% & 10.8\% & 41.8\% \\
\midrule
All & 47.4\% & 8.5\% & 44.1\% & 45.7\% & 12.5\% & 41.8\% \\
\bottomrule
\end{tabular}}
\vspace{-.05in}
\label{tab:user_study}
\end{table}

\subsection{Text-to-Video and Image-to-Video Generation}
\label{Video gen}
\noindent\textbf{Implementation Details.}
For video generation, we employ three pre-trained models: Wan2.1-T2V-1.3B, Wan2.1-T2V-14B, and Wan2.1-I2V-14B. Each model is adapted from the original Wan2.1-VAE to our DC-AE-V (\secref{sec:application}). Detailed hyperparameters are provided in \secref{sec:hyper_params}.

\vspace{0.1cm}
\noindent\textbf{Dataset and Evaluation Protocol.}
For training, we collect 257K synthetic videos using FusionX\footnote{\url{https://civitai.green/models/1651125/wan2114bfusionx}}
, a high-quality variant of Wan2.1-T2V-14B.
Following common practice, we use VBench~\cite{huang2023vbench} for text-to-video evaluation and VBench-I2V~\cite{huang2025vbench++} for image-to-video evaluation. We adopt the extended prompt sets\footnote{\url{https://github.com/Vchitect/VBench/tree/master/prompts}}
 provided by the VBench team and conduct all evaluations at the same resolution to ensure fair, apples-to-apples comparisons.
To rule out the influence of training data, we additionally perform fine-tuning to pre-trained models (referred to as \textbf{Tuned} in \tabref{tab:video_720}) on our dataset using the same hyperparameters.

\begin{table*}[t]
\caption{
\textbf{Results on Video Generation.} $^{\dag}$Native Wan2.1-T2V-1.3B is limited to 480$\times$832 resolution, so we fine-tune it on our dataset to support 720$\times$1280 generation.}
\vspace{-10pt}
\small\centering\setlength{\tabcolsep}{3pt}
\resizebox{\linewidth}{!}{
\begin{tabular}{g | g | g | g | g | g g}
\toprule
\multicolumn{7}{l}{\textbf{Video Generation Results on VBench 720$\times$1280}} \\
\midrule
\rowcolor{white} & &  & Latency & \multicolumn{3}{c}{Score $\uparrow$} \\
\cmidrule{5-7}
\rowcolor{white} \multirow{-2}{*}{Video Diffusion Model} & \multirow{-2}{*}{Video Autoencoder} & \multirow{-2}{*}{Version} & (min) $\downarrow$ & Overall & Quality & Semantic/I2V \\
\midrule
\rowcolor{white} MAGI-1~\cite{teng2025magi} & - & Pre-trained T2V & 21.22 & 79.18 & 82.04 & 67.74 \\
\rowcolor{white} Step-Video~\cite{ma2025step} & - & Pre-trained T2V & 13.16 & 81.83 & 84.46 & 71.28 \\
\rowcolor{white} CogVideoX1.5~\cite{yang2024cogvideox} & - & Pre-trained T2V & 6.73 & 82.17 & 82.78 & 79.76 \\
\rowcolor{white} Skyreels-V2~\cite{chen2025skyreels} & - & Pre-trained T2V & 9.48 & 82.67 & 84.70 & 74.53 \\
\rowcolor{white} HunyuanVideo~\cite{kong2024hunyuanvideo} & - & Pre-trained T2V & 30.35 & 83.24 & 85.09 & 75.82 \\
\rowcolor{white} OpenSora-2.0~\cite{peng2025open} & - & Pre-trained T2V & 32.83 & 84.34 & 85.40 & 80.12 \\
\midrule
\midrule
\rowcolor{white} CogVideoX-5b-I2V~\cite{yang2024cogvideox} & - & Pre-trained I2V & 6.72 & 86.70 & 78.61 & 94.79 \\
\rowcolor{white} HunyuanVideo-I2V~\cite{kong2024hunyuanvideo} & - & Pre-trained I2V & 30.39 & 86.82 & 78.54 & 95.10 \\
\rowcolor{white} Step-Video-TI2V~\cite{ma2025step} & - & Pre-trained I2V & 13.18 & 88.36 & 81.22 & 95.50 \\
\rowcolor{white} MAGI-1~\cite{teng2025magi} & - & Pre-trained I2V & 21.25 & 89.28 & 82.44 & 96.12 \\
\midrule
\midrule
\rowcolor{white} Wan2.1-T2V-1.3B$^{\dag}$~\cite{wan2025} & Wan2.1-VAE-f8c16 & Tuned & 5.76 & 83.38 & 85.67 & 74.22 \\
DC-Gen-Wan2.1-T2V-1.3B & DC-AE-V-f32c32 & DC-Gen & 0.70 & 83.87 & 86.38 & 73.80 \\
\midrule
\midrule
\rowcolor{white} Wan2.1-T2V-14B~\cite{wan2025} & Wan2.1-VAE-f8c16 & Pre-trained T2V & 27.52 & 83.73 & 85.77 & 75.58 \\
\rowcolor{white} Wan2.1-T2V-14B~\cite{wan2025} & Wan2.1-VAE-f8c16 & Tuned & 27.52 & 83.94 & 86.18 & 74.97 \\
DC-Gen-Wan2.1-T2V-14B & DC-AE-V-f32c32 & DC-Gen & 3.58 & 83.97 & 86.00 & 75.83 \\
\midrule
\midrule
\rowcolor{white} Wan2.1-I2V-14B~\cite{wan2025} & Wan2.1-VAE-f8c16 & Pre-trained I2V & 27.88 & 86.86 & 80.83 & 92.90 \\
\rowcolor{white} Wan2.1-I2V-14B~\cite{wan2025} & Wan2.1-VAE-f8c16 & Tuned & 27.88 & 86.72 & 81.19 & 92.25 \\
DC-Gen-Wan2.1-I2V-14B & DC-AE-V-f32c32 & DC-Gen & 3.67 & 87.13 & 81.20 & 93.06 \\
\bottomrule
\end{tabular}}

\label{tab:video_720}
\end{table*}

\begin{figure*}[!tb]
    \centering
    \animategraphics[width=\linewidth]{8}{figures/videosrc/video_gen_samples_main_text/}{00000}{00060}
      \caption{Visual Comparison of DC-Gen-Wan2.1-14B and Wan2.1-14B.}
\label{fig:video_gen_samples_main_text}
\end{figure*}

\vspace{0.1cm}
\noindent\textbf{Experimental Results.}
Table~\ref{tab:video_720} compares DC-Gen with leading T2V and I2V diffusion models on VBench at 720$\times$1280 resolution. 
Compared with the pre-trained Wan2.1 models, DC-Gen-Wan2.1 achieves comparable quantitative results while being significantly faster. For example, DC-Gen-Wan2.1-T2V-14B reduces latency by 7.7$\times$. Compared with other pre-trained diffusion models, DC-Gen-Wan2.1 series achieves both relatively high VBench scores and the lowest latency. As demonstrated in the video samples in \figref{fig:video_gen_samples_main_text}, as well as the webpage in the supplementary material, DC-Gen achieves comparable visual quality to the pre-trained model with faster inference speed.

Similar to text-to-image generation, we also conducted human evaluation for both T2V and I2V generation, with results presented in \secref{sec:human_study_videogen}.

\subsection{Image Editing}
\label{Image Editing}

\noindent\textbf{Implementation Details.}
We primarily adopt Qwen-Image-Edit~\cite{wu2025qwenimagetechnicalreport} as the pre-trained model. To ensure effective alignment, we fine-tune the pre-trained model on a high-quality dataset before applying our DC-Gen pipeline. 

\noindent\textbf{Dataset and Evaluation Protocol.} We use GEdit~\cite{liu2025step1x-edit} as the benchmark. GEdit utilizes pre-trained VLM to measure the fidelity of image editing to the given instruction (SC score), the quality of the output image (PQ Score), and the overall behavior of editing (O Score, which is the geometric average of SC Score and PQ Score), while supporting both English and Chinese. We use Qwen-2.5 as the discriminative VLM backbone~\cite{Qwen2.5-VL}. Our training dataset consists of 201K selected editing pairs from Pico-Banana~\cite{qian2025picobanana400klargescaledatasettextguided}. For fair comparison, we also fine-tuned the pre-trained model on this dataset (referred to as \textbf{Tuned} in \tabref{tab:image_edit_1024}).

\begin{table*}[htbp]
\vspace{-5pt}
\caption{
\textbf{Image Editing Results.} $^{\dag}$Native FLUX.1-Kontext only supports English.}
\vspace{-10pt}
\small\centering\setlength{\tabcolsep}{3pt}
\resizebox{0.95\linewidth}{!}{
\begin{tabular}{g | g | g | g | g  g  g | g  g g}
\toprule
\multicolumn{10}{l}{\textbf{Image Editing Results on GEdit 1K}} \\
\midrule
\rowcolor{white} & & & Latency & \multicolumn{3}{c|}{GEdit-English $\uparrow$} & \multicolumn{3}{c}{GEdit-Chinese $\uparrow$} \\
\cmidrule{5-10}
\rowcolor{white} \multirow{-2}{*}{Model} & \multirow{-2}{*}{Version} & \multirow{-2}{*}{Autoencoder}  & (s) $\downarrow$ & SC & PQ & O & SC & PQ & O \\
\midrule
\rowcolor{white} FLUX.1-Kontext$^{\dag}$~\cite{labs2025flux1kontextflowmatching} & Pre-trained & FLUX-VAE-f8c16 & 14.67 & 6.15 & 7.17 & 5.77 & - & - & - \\
\rowcolor{white} Step1X-Edit~\cite{liu2025step1x-edit} & Pre-trained & FLUX-VAE-f8c16 & 49.75 & 6.16 & 7.01 & 5.73 & 5.43 & 7.27 & 5.45 \\
\midrule
\midrule
\rowcolor{white} Qwen-Image-Edit~\cite{wu2025qwenimagetechnicalreport} & Pre-trained & Qwen-VAE-f8c16 & 57.63 & 6.10 & 7.36 & 5.81 & 5.58 & 7.41 & 5.59 \\
\rowcolor{white} Qwen-Image-Edit~\cite{wu2025qwenimagetechnicalreport} & Tuned & Qwen-VAE-f8c16 & 57.63 & 7.64 & 7.36 & 7.23 & 7.37 & 7.32 & 7.02 \\
DC-Gen-Qwen-Image-Edit & DC-Gen & DC-AE-f32c32 & 11.96 & 7.71 & 7.12 & 7.20 & 7.78 & 7.15 & 7.32 \\
\bottomrule
\end{tabular}}
\label{tab:image_edit_1024}
\end{table*}

\begin{figure*}[h!]
    \centering
    \vspace{-.05in}
    \includegraphics[width=\linewidth]{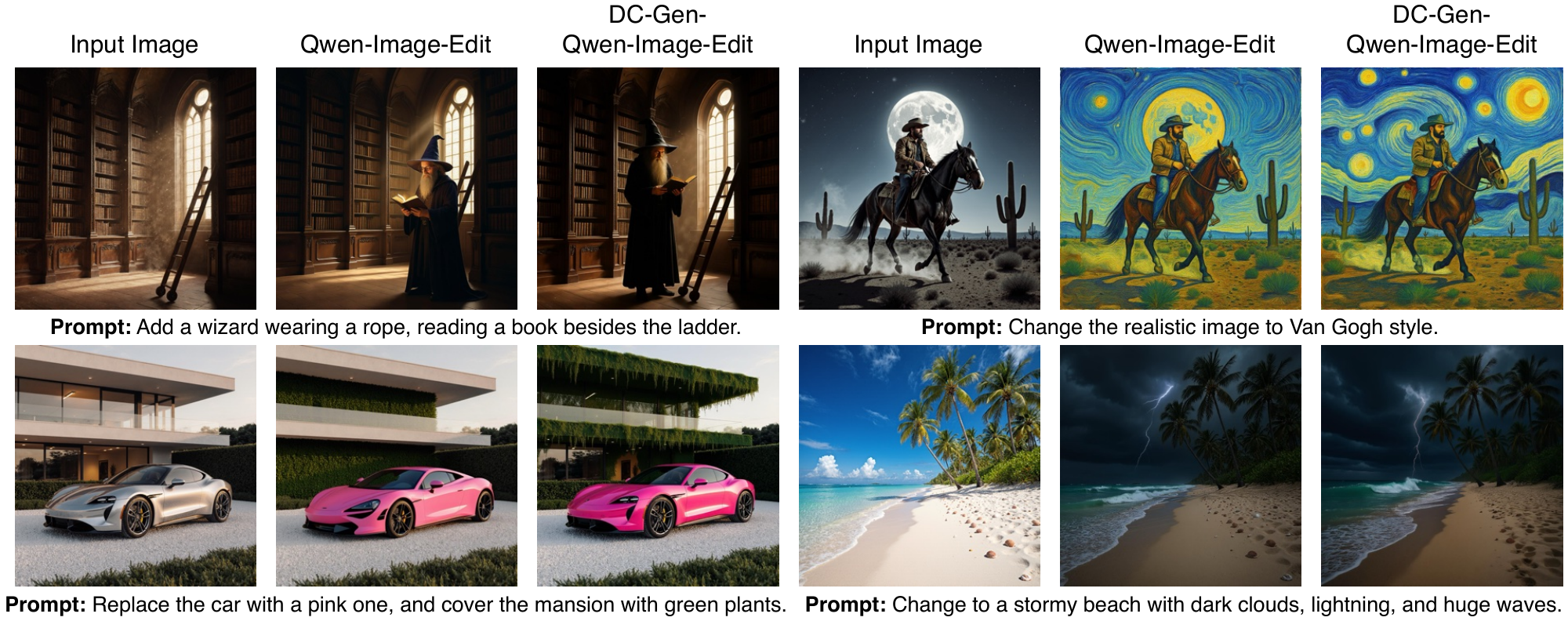}
    \vspace{-.25in}
    \caption{
   \textbf{Visual Results on Image Editting.} }
\label{fig:image_edit_samples_main_text}
\vspace{-.1in}
\end{figure*}

\noindent \textbf{Experimental Results.} \tabref{tab:image_edit_1024} compares DC-Gen-Qwen-Image-Edit with leading open-source image editing models. Consistent with the results in image and video generation, DC-Gen achieves comparable editing performance while reducing latency by 4.8$\times$. \figref{fig:image_edit_samples_main_text} and \secref{sec:human_study_image_edit} further shows the qualitative comparison. Human evaluation results are provided in \secref{sec:human_study_image_edit}.

\section{Conclusion}
\label{sec:conclusion}
We present DC-Gen, a novel framework that accelerates pre-trained diffusion models by adapting them to deeply compressed latent spaces. Our approach addresses the challenge of quality degradation relative to the pre-trained models through a novel embedding alignment procedure, followed by end-to-end fine-tuning to preserve the model’s generative capabilities after switching latent spaces. We validate DC-Gen across image generation, video generation, and image editing tasks. The resulting models match the performance of their pre-trained counterparts while achieving substantial efficiency gains. In Sec.~\ref{sec:orthogonal}, we also demonstrated that DC-Gen can be well-combined with other orthogonal acceleration methods. 

\clearpage

\appendix

\section{Corrected Flow Matching Objective for Guidance-Distilled Models}
\label{sec:guidance_distilled}

\begin{figure}[t]
    \centering
    \begin{subfigure}[!tb]{0.53\linewidth}
        \centering
        \includegraphics[width=\linewidth]{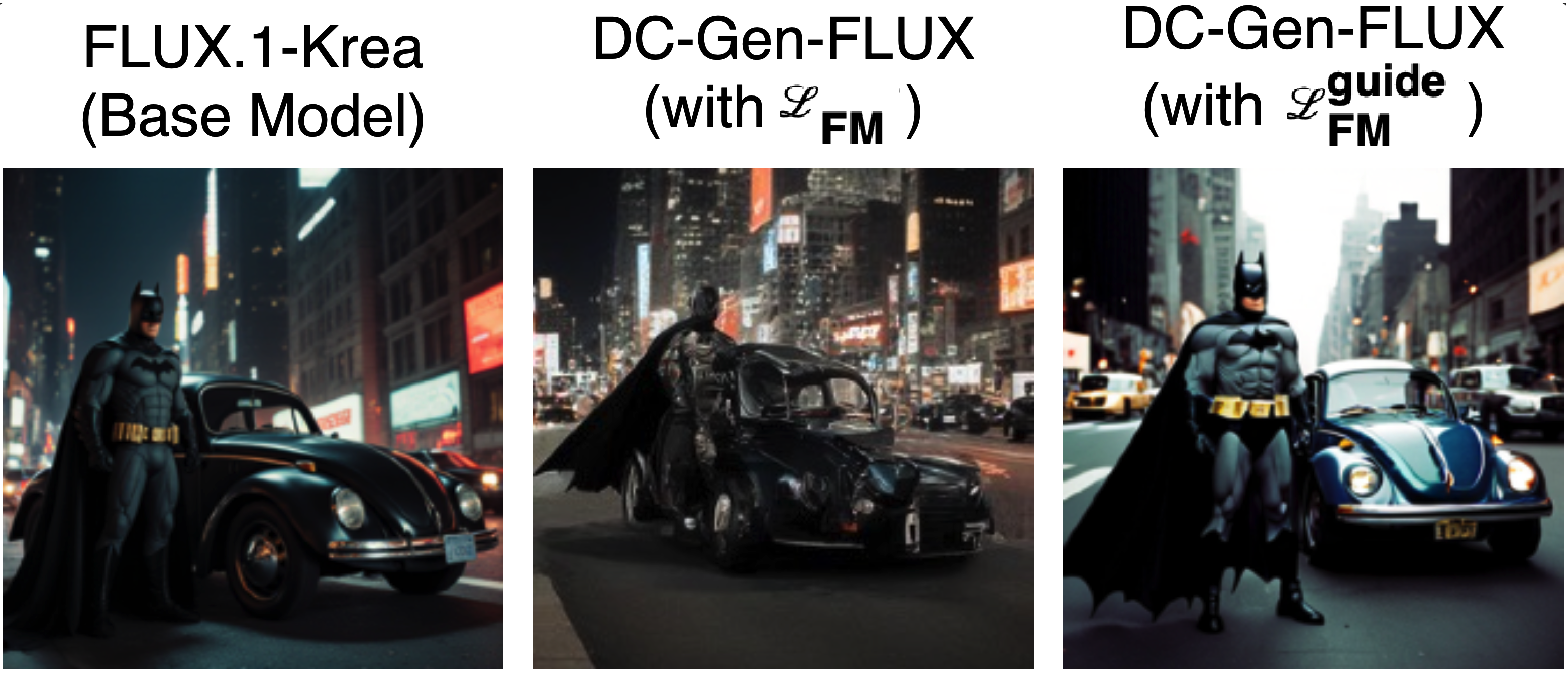}
        \vspace{-.2in}
        \caption{Visual comparison of training objectives.}
        \label{fig:ablation_guidetrain_visual}
    \end{subfigure}
    \hfill
    \begin{subtable}[!tb]{0.45\linewidth}
        \caption{Quantitative comparison of training objectives.}
        \centering\renewcommand{\arraystretch}{1.7}
        \resizebox{\linewidth}{!}{
        \begin{tabular}{c|c}
            \toprule
            \rowcolor{white}\textbf{Method}  & \textbf{CLIP Score}$\uparrow$ \\
            \midrule
            \rowcolor{white}Train with $\mathcal{L}_{\text{FM}}$ (\equref{eq:fm})  & 26.87 \\
            \rowcolor{cyan!10}Train with $\mathcal{L}_{\text{FM}}^{\text{guide}}$ (\equref{eq:fm_guide}) & \textbf{27.38} \\
            \bottomrule
        \end{tabular}}
        \vspace{-.05in}
        \label{tab:ablation_guidetrain_metric}
    \end{subtable}
    \vspace{-.1in}
    \caption{\textbf{Ablation of Training Objectives for Guidance-Distilled Models.} $\mathcal{L}_{\text{FM}}^{\text{guide}}$ (\equref{eq:fm_guide}) corrects the velocity estimation and demonstrates a large improvement in visual quality. The metric is reported MJHQ-30K 512$\times$512. }
    \label{fig:ablation_guidetrain}
    \vspace{-.2in}
\end{figure}

Our pre-trained text-to-image models, FLUX.1-Krea-12B~\cite{flux1kreadev2025} and Z-Image-Turbo~\cite{imageteam2025zimageefficientimagegeneration}, are guidance-distilled models. Adopting the standard flow-matching objective
\begin{equation}
    \mathcal{L}_{\text{FM}}(\theta) = \mathbb{E}_{\mathbf{z}_0, \boldsymbol{\epsilon}, t, w} \left[ \left\| \mathbf{v}_{\theta}(\mathbf{z}_t, \mathbf c, t, w) - \mathbf{v}_t \right\|_2^2 \right] \label{eq:fm}
\end{equation}
for training the guidance-distilled models will result in a biased estimation of the velocity, as exemplified by the results in \figref{fig:ablation_guidetrain} (with $\mathcal{L}_{\text{FM}}$). To solve this problem, we analyze and provide a corrected training objective below.

\myPara{Revisiting Guidance Distillation.} The core issue stems from how the guidance-distilled model, denoted as $\mathbf{v}_{\eta}$, is trained. The goal of guidance distillation~\cite{meng2023distillation} is to train $\mathbf{v}_{\eta}$ to mimic the output of classifier-free guidance (CFG):
\begin{equation}
    \mathcal{L}_{\text{distill}}(\eta)=\mathbb{E}_{\mathbf{z}_0, \boldsymbol{\epsilon}, t, w} \left[\left\| \mathbf{v}_{\eta}(\mathbf{z}_t, \mathbf c, t, w) - \mathbf{v}_{\theta}^w(\mathbf{z}_t, \mathbf c, t) \right\|_2^2 \right].\label{eq:distill}
\end{equation}
Here, $w\in [w_{\text{min}}, w_{\text{max}}]$ represents the guidance scale, and
\begin{equation}
\mathbf{v}_{\theta}^w(\mathbf{z}_t, \mathbf c, t) = (1+w)\mathbf{v}_{\theta}(\mathbf{z}_t, \mathbf c, t) - w\mathbf{v}_{\theta}(\mathbf{z}_t, \hat{\mathbf c}, t)\label{eq:cfg}
\end{equation}
is the CFG objective, where $\mathbf c$ is the text condition, and $\hat{\mathbf c}$ is the negative text condition.

Since the guidance-distilled model $\mathbf{v}_{\eta}$ is trained to produce this CFG output, simply applying the standard flow-matching objective will be incorrect.

\myPara{Correcting the Velocity Estimation.}
To address this, we must derive the ``raw'' velocity $\mathbf{v}_\theta$ from the distilled model $\mathbf{v}_\eta$. We can do so by recognizing that the unconditional velocity, $\mathbf{v}_\theta(\mathbf{x}_t, \hat{\mathbf c}, t)$, can be approximated by the distilled model's output when the condition is empty, \ie,  
\begin{align}
\mathbf{v}_{\eta}(\mathbf{z}_t, \hat{\mathbf c}, t, w)&\approx \mathbf{v}_{\theta}^w(\mathbf{z}_t, \hat{\mathbf c}, t) && \text{from }(\equref{eq:distill}) \nonumber \\
&=(1+w)\mathbf{v}_{\theta}(\mathbf{z}_t, \hat{\mathbf c}, t) - w\mathbf{v}_{\theta}(\mathbf{z}_t, \hat{\mathbf c}, t) && \text{from }(\equref{eq:cfg})\nonumber \\
&=\mathbf{v}_\theta(\mathbf{z}_t, \hat{\mathbf c}, t).\label{eq:unconditional_velocity}
\end{align}
By treating the distilled model's output as the CFG output, we have:
\begin{align}
\mathbf{v}_{\eta}(\mathbf{z}_t, \mathbf c, t, w)&\approx \mathbf{v}_{\theta}^w(\mathbf{z}_t, \mathbf c, t) && \text{from }(\equref{eq:distill}) \nonumber \\
&=(1+w)\mathbf{v}_{\theta}(\mathbf{z}_t, \mathbf c, t) - w\mathbf{v}_{\theta}(\mathbf{z}_t, \hat{\mathbf c}, t) && \text{from }(\equref{eq:cfg}) \nonumber \\
&\approx (1+w)\mathbf{v}_{\theta}(\mathbf{z}_t, \mathbf c, t) - w\mathbf{v}_{\eta}(\mathbf{z}_t, \hat{\mathbf c}, t, w). && \text{from }(\equref{eq:unconditional_velocity})
\end{align}
To solve for the desired ``raw'' velocity $\mathbf{v}_{\theta}(\mathbf{z}_t, \mathbf c, t)$, we can algebraically rearrange this equation:
\begin{equation}
    \mathbf{v}_{\theta}(\mathbf{z}_t, \mathbf c, t) \approx \frac{1}{1+w}\left(\mathbf{v}_{\eta}(\mathbf{z}_t, \mathbf c, t, w) + w\mathbf{v}_{\eta}(\mathbf{z}_t, \hat{\mathbf c}, t, w)\right),
\end{equation}
which gives us the corrected velocity estimation. The objective thus becomes:
\begin{equation}
\mathcal{L}_{\text{FM}}^{\text{guide}}
=
\mathbb{E}_{\mathbf{z}_0, \boldsymbol{\epsilon}, t, w}
\left[
\left\|
\frac{1}{1+w}
\left(
\mathbf{v}_{\eta}(\mathbf{z}_t, \mathbf c, t, w)
+
w\mathbf{v}_{\eta}(\mathbf{z}_t, \hat{\mathbf c}, t, w)
\right)
-
\mathbf{v}_t
\right\|_2^2
\right].
\label{eq:fm_guide}
\end{equation}
With the $\mathcal{L}_{\text{FM}}^{\text{guide}}$, we are able to directly train the guidance-distilled model, as shown in \figref{fig:ablation_guidetrain}.

\section{Additional Ablation Studies}
\label{sec:intro}

\subsection{Embedding Alignment}
\label{sec:appendix_embedding_alignment}
In \secref{sec:discussion}, we experiment on various diffusion models and tasks to demonstrate the significance of \textbf{embedding alignment}, which enables the new model in the deeply compressed latent space to fully recover the pre-trained model's generative capability. In this section, we conduct further ablation studies on the two stages of embedding alignment: the \textbf{patch embedder alignment} and the \textbf{output layer alignment}. FLUX.1-Krea~\cite{flux1kreadev2025} for text-to-image and Wan2.1-T2V-1.3B~\cite{wan2025} for text-to-video serve as the pre-trained models.

\begin{figure}[htbp]
    \centering
    \vspace{-.1in}
    \includegraphics[width=0.7\linewidth]{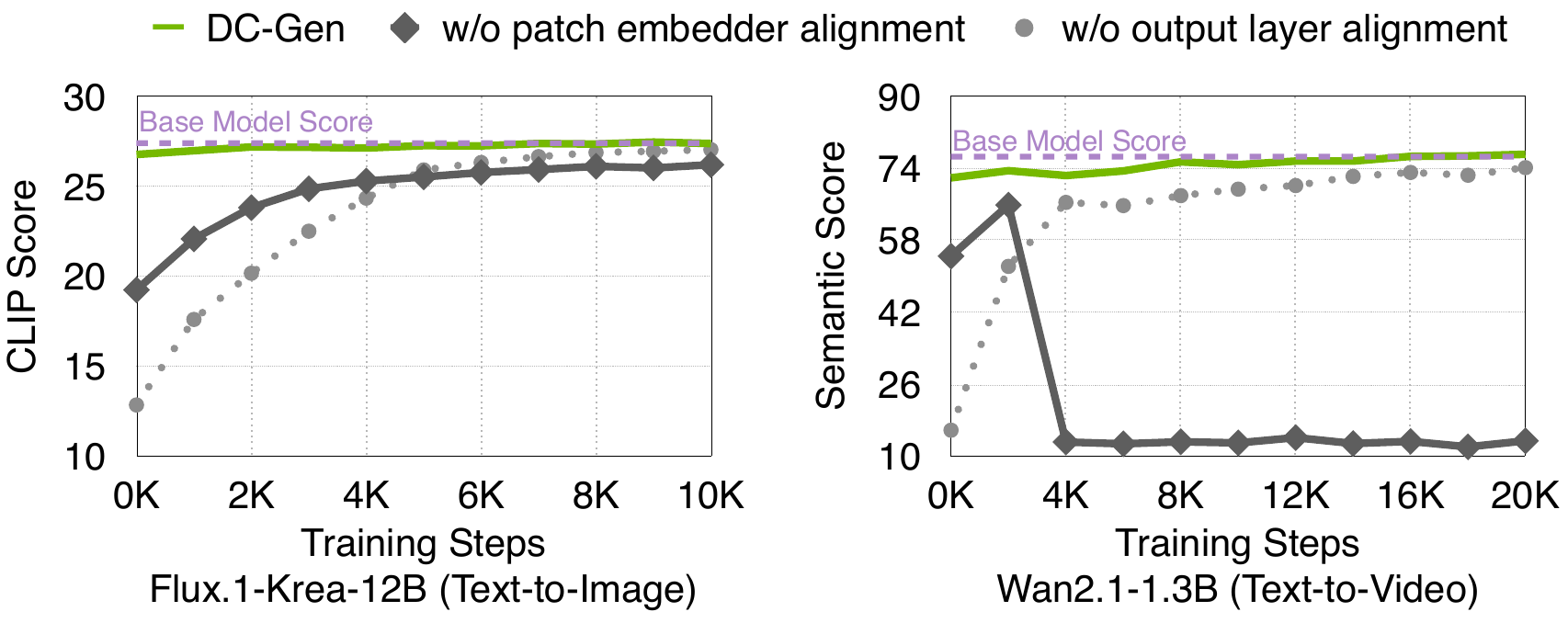}
    \vspace{-.08in}
    \caption{
         \textbf{Training Curve Comparison of Ablations on Embedding Alignment.}}
    \label{fig:appendix_ablation_alignment_metrics}
    \vspace{-.1in}
\end{figure}

\begin{figure}[h!]
    \centering
    \includegraphics[width=0.9\linewidth]{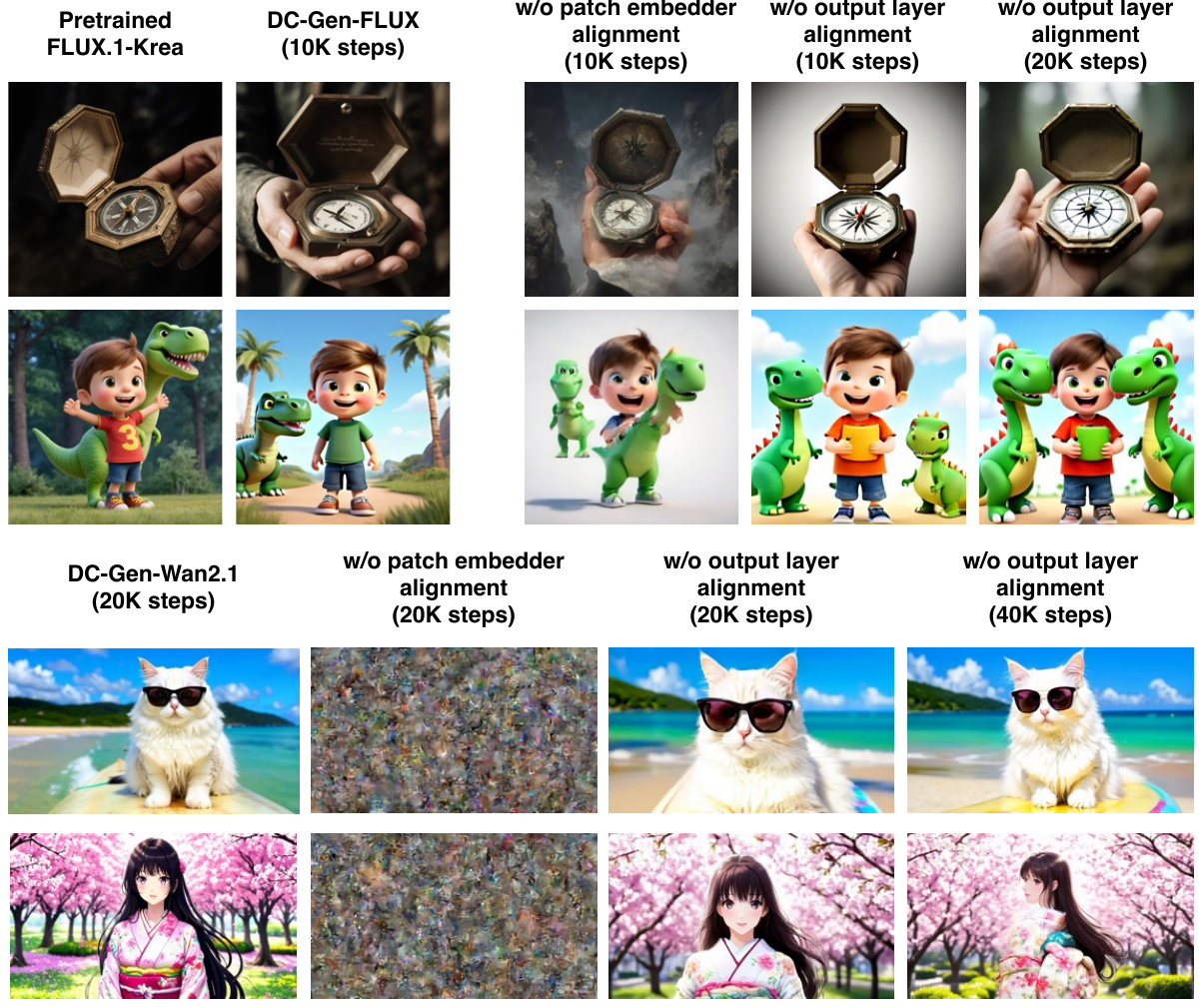}
    \vspace{-.1in}
    \caption{
         \textbf{Visual Comparison of Ablations on Embedding Alignment.} }
    \label{fig:appendix_ablation_alignment_vis}
    \vspace{-.1in}
\end{figure}

As shown in \figref{fig:appendix_ablation_alignment_metrics} and \figref{fig:appendix_ablation_alignment_vis}, without patch embedder alignment, even if we jointly fine-tune the patch embedder and output layer before the end-to-end fine-tuning, there is no significant difference in visual quality and metrics compared with performing no embedding alignment at all. For FLUX.1-Krea, the model generates images with blurred details and artifacts, while Wan2.1-T2V-1.3B still suffers from training instability, with generated videos finally degrading to pure noise. By contrast, without output layer alignment, the convergence speed for the end-to-end fine-tuning stage slows down, taking at least double steps to achieve similar visual quality and quantitative metrics.

We further compare DC-Gen with two additional adaptation strategies, which might stabilize the end-to-end fine-tuning process to some extent:

\begin{itemize}[leftmargin=*]
    \item \textbf{Layer-Wise Learning Rate Decay.} Instead of using a uniform learning rate, we apply a linear layer-wise learning rate decay, where earlier layers receive smaller learning rates and later layers receive larger ones, while keeping the average learning rate unchanged. The ratio between the smallest and largest learning rates is 1:3.
    \item \textbf{Gradually Introducing LoRA Modules.} In this baseline, we use a linear warmup schedule for the LoRA alpha value, starting at 0 and increasing linearly to the target value.
\end{itemize}

\begin{table*}[ht]
\caption{\textbf{Quantitative Comparison with Additional Adaptation Strategies.}}
\small\centering\setlength{\tabcolsep}{3pt}
\resizebox{0.7\linewidth}{!}{
\begin{tabular}{g | g | g |g g }
\toprule
\multicolumn{5}{l}{\textbf{Text-to-Image Generation Results on MJHQ-30K 512$\times$512}} \\
\midrule
\rowcolor{white} & & &  \\
\rowcolor{white} \multirow{-2}{*}{\textbf{Model}} & \multirow{-2}{*}{\textbf{Autoencoder}} & \multirow{-2}{*}{\textbf{Method}} & \multirow{-2}{*}{\textbf{FID} $\downarrow$}  & \multirow{-2}{*}{\textbf{CLIP Score} $\uparrow$}  \\
\midrule
\rowcolor{white} & & Layer-Wise LR Decay & 17.83 & 25.77 \\
\rowcolor{white} & & Gradual LoRA & 15.90 & 26.52 \\
\cellcolor{white} \multirow{-3}{*}{FLUX.1-Krea} & \cellcolor{white} \multirow{-3}{*}{DC-AE-f32c32} & DC-Gen & \textbf{13.30} & \textbf{27.38} \\
\bottomrule
\end{tabular}}
\label{tab:appendix_ablation_embedding_alignment}
\end{table*}

\begin{figure}[h!]
    \centering
    \includegraphics[width=0.72\linewidth]{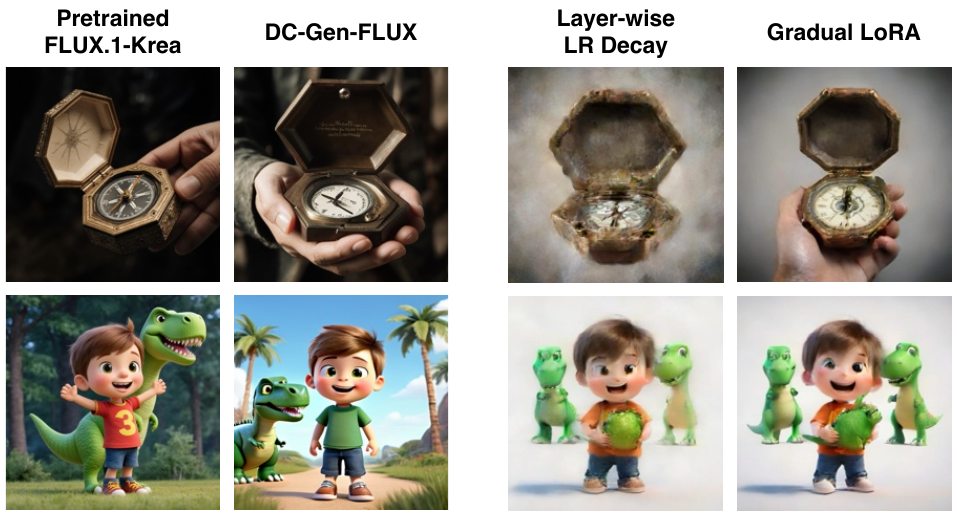}
    \vspace{-.1in}
    \caption{
         \textbf{Visual Comparison of DC-Gen with Other Techniques.} }
    \label{fig:appendix_ablation_stabilize}
    \vspace{-.1in}
\end{figure}

As shown in \tabref{tab:appendix_ablation_embedding_alignment} and \figref{fig:appendix_ablation_stabilize}, we conduct experiments on FLUX.1-Krea~\cite{flux1kreadev2025}, yet none of these techniques achieve comparable visual quality against the pre-trained model and DC-Gen.

Apart from embedding alignment, another idea is to perform latent space alignment at the autoencoder level. Specifically, we fine-tune the deep compression autoencoder using an additional alignment loss that encourages its latent space to match that of the original autoencoder. We experiment on FLUX.1-Krea at 512px resolution, adapting from FLUX-VAE-f8c16 to DC-AE-f32c32~\cite{chen2024deep}. We try three different alignment loss ratios—0.1, 1.0, and 10.0. As shown in \tabref{tab:appendix_ablation_vae_alignment}, DC-Gen with embedding alignment achieves substantially better results compared to Autoencoder Latent Space Alignment.

\begin{table}[h]
\caption{\textbf{Comparison with Autoencoder Latent Space Alignment.} Base model is FLUX.1-Krea-12B. }
\small\centering\setlength{\tabcolsep}{10pt}
\resizebox{0.9\linewidth}{!}{
\begin{tabular}{g | g | g g }
\toprule
\multicolumn{4}{l}{\textbf{Text-to-Image Generation Results on MJHQ-30K 512$\times$512}} \\
\midrule
\rowcolor{white} \textbf{Method} & \textbf{AE Alignment Ratio} & \textbf{gFID} $\downarrow$ & \textbf{CLIP-Score} $\uparrow$ \\
\midrule
\rowcolor{white}  & 0.1 & 15.39 & 26.53 \\
\rowcolor{white} Autoencoder Latent Space Alignment & 1.0 & 15.27 & 26.36 \\
\rowcolor{white} & 10.0 & 21.76 & 23.20 \\
\midrule
DC-Gen & - & \textbf{13.30} & \textbf{27.38} \\
\bottomrule
\end{tabular}}
\label{tab:appendix_ablation_vae_alignment}
\end{table}

To sum up, we conclude that DC-Gen with the embedding alignment is, to the best of our knowledge, the only technique currently available to achieve effective deeply compressed latent space adaptation. Its two phases—the patch embedder alignment and the output layer alignment—are both indispensable for maintaining training stability and accelerating the convergence of end-to-end fine-tuning.

\subsection{LoRA Fine-Tuning}

As discussed in the Sec. 3.3 of the main paper, since the model after embedding alignment has already been synchronized with the new latent space and can generate visual samples with reasonable semantics, we use LoRA for the end-to-end fine-tuning stage. Here we conduct ablation studies on LoRA fine-tuning.

\begin{table}[htbp]
    \centering
    \caption{\textbf{Ablation of LoRA Fine-Tuning.}}
    \begin{subtable}[!tb]{0.475\linewidth}
        \caption{Quantitative comparison of end-to-end fine-tuning techniques on FLUX.1-Krea.}
        \centering\renewcommand{\arraystretch}{1.7}
        \resizebox{\linewidth}{!}{
        \begin{tabular}{l|g|g|g}
            \toprule
            \multicolumn{4}{l}{\textbf{Text-to-Image Generation Results on MJHQ-30K 512$\times$512}} \\
            \midrule
            \rowcolor{white} Text-to-Image &  & Trainable &  \\
            \rowcolor{white}Diffusion Model & \multirow{-2}{*}{Method} &  Params (M) &  \multirow{-2}{*}{CLIP Score$\uparrow$} \\
            \midrule
            \rowcolor{white} & w/o LoRA  & 11901.21 & 26.85 \\
            \cellcolor{white} \multirow{-3}{*}{FLUX.1-Krea~\cite{flux1kreadev2025}} & DC-Gen & 1046.74 & \textbf{27.38} \\
            \bottomrule
        \end{tabular}}
        \vspace{-.05in}
        \label{tab:ablation_guidetrain_metric}
    \end{subtable}
    \hfill
    \begin{subtable}[!tb]{0.505\linewidth}
        \caption{Quantitative comparison of end-to-end fine-tuning techniques on Wan2.1-T2V-1.3B.}
        \centering\renewcommand{\arraystretch}{1.7}
        \resizebox{\linewidth}{!}{
        \begin{tabular}{l | g | g | g  g g}
            \toprule
            \multicolumn{6}{l}{\textbf{Text-to-Video Generation Results on VBench 480$\times$832}} \\
            \midrule
            \rowcolor{white} Text-to-Video & & Trainable & \multicolumn{3}{c}{Score $\uparrow$}  \\
            \cmidrule{4-6}
            \rowcolor{white} Diffusion Model & \multirow{-2}{*}{Method} & Params (M) & Overall & Quality & Semantic \\
            \midrule
            \rowcolor{white} & w/o LoRA & 1418.90 & 79.81 & 84.02 & 62.98 \\
            \cmidrule{2-6}
            \multirow{-2}{*}{Wan2.1-1.3B~\cite{wan2025}} & DC-Gen & 350.37 & \textbf{83.48} & \textbf{85.08} & \textbf{77.12} \\
            \bottomrule
        \end{tabular}}
        \vspace{-.05in}
        \label{tab:ablation_guidetrain_metric}
    \end{subtable}
    \vspace{-.1in}
    \label{tab:appendix_ablation_end2end}
\end{table}

\begin{figure}[htbp]
    \centering
    \vspace{-.1in}
    \includegraphics[width=0.7\linewidth]{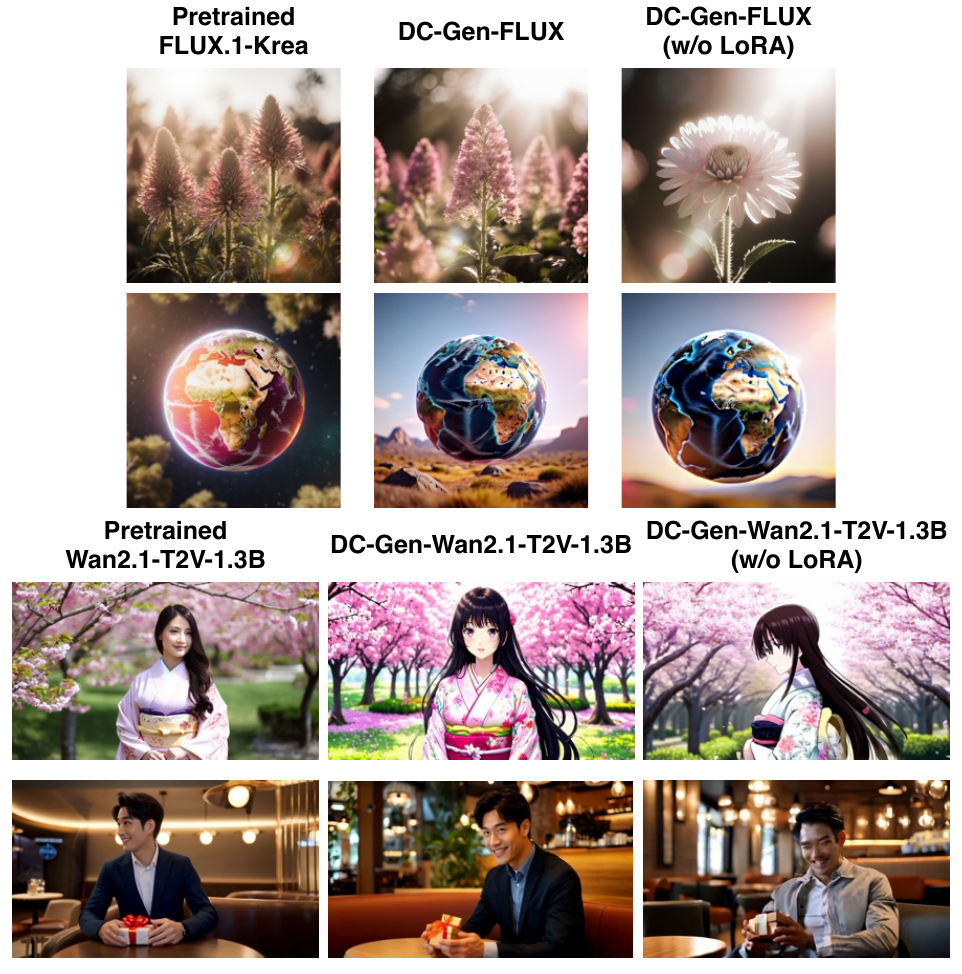}
    \vspace{-.1in}
    \caption{
         \textbf{Visual Comparison of Ablations for LoRA Fine-Tuning.} }
    \vspace{-.1in}
    \label{fig:appendix_ablation_end2end}
    \vspace{-.05in}
\end{figure}

As demonstrated in \tabref{tab:appendix_ablation_end2end} and \figref{fig:appendix_ablation_end2end}, LoRA fine-tuning not only saves memory cost during training by significantly reducing trainable parameters, but also yields better visual quality, which might be attributed to its improved preservation of the pre-trained model’s generative capabilities.

\section{Visualization Samples}

\subsection{Text-to-Image Generation}
\label{sec:appendix_t2i_samples}
\begin{figure}[htbp]
    \centering
    \includegraphics[width=0.95\linewidth]{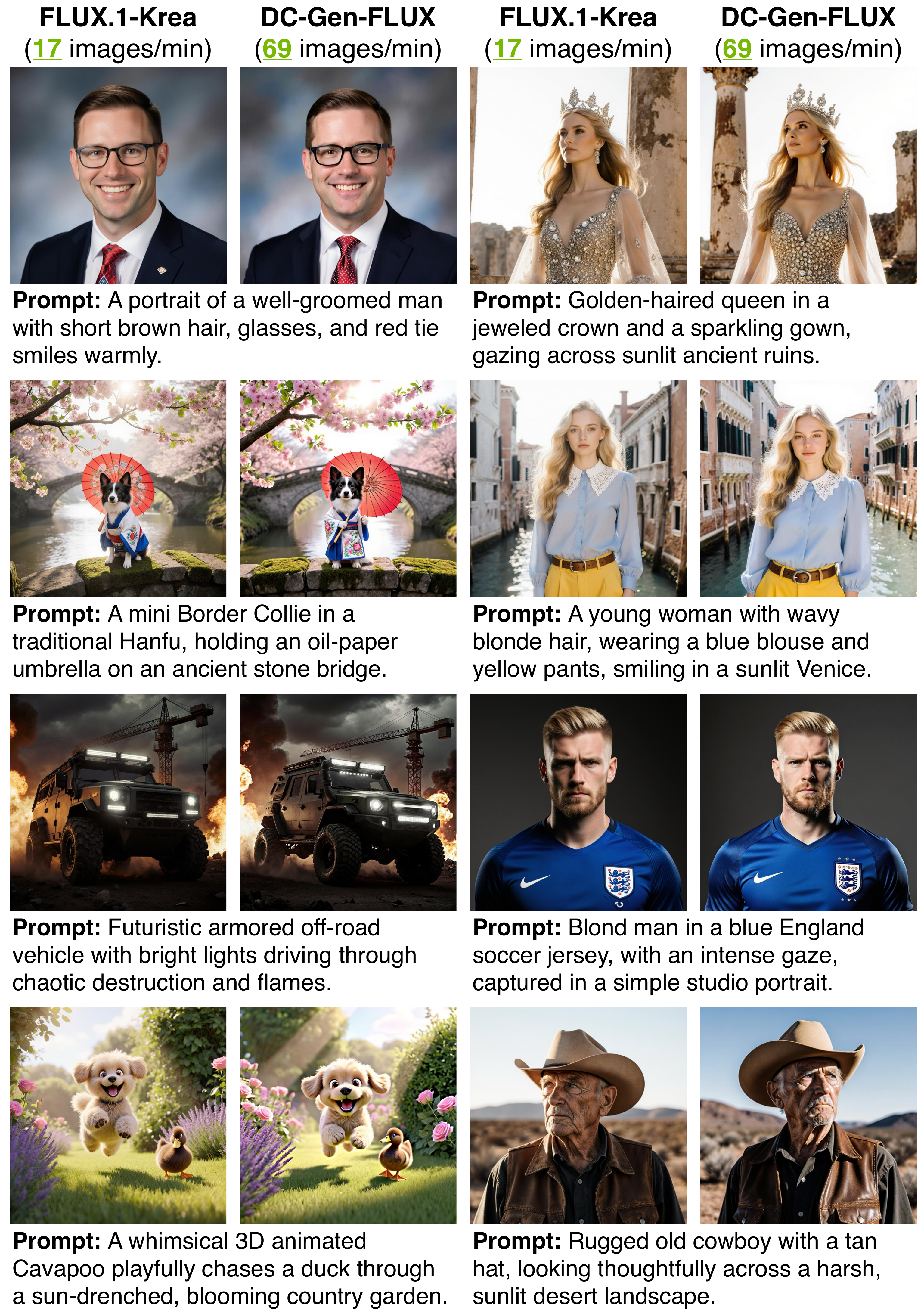}
    \caption{\textbf{Samples of FLUX.1-Krea and DC-Gen-FLUX on 1024$\times$1024 Resolution.} DC-Gen-FLUX preserves the generation quality and superior realism of FLUX.1-Krea, while achieving an approximately 4$\times$ throughput improvement.
    }
    \label{fig:appendix_flux_sample1k}
\end{figure}

\begin{figure}[htbp]
    \centering
    \includegraphics[width=0.95\linewidth]{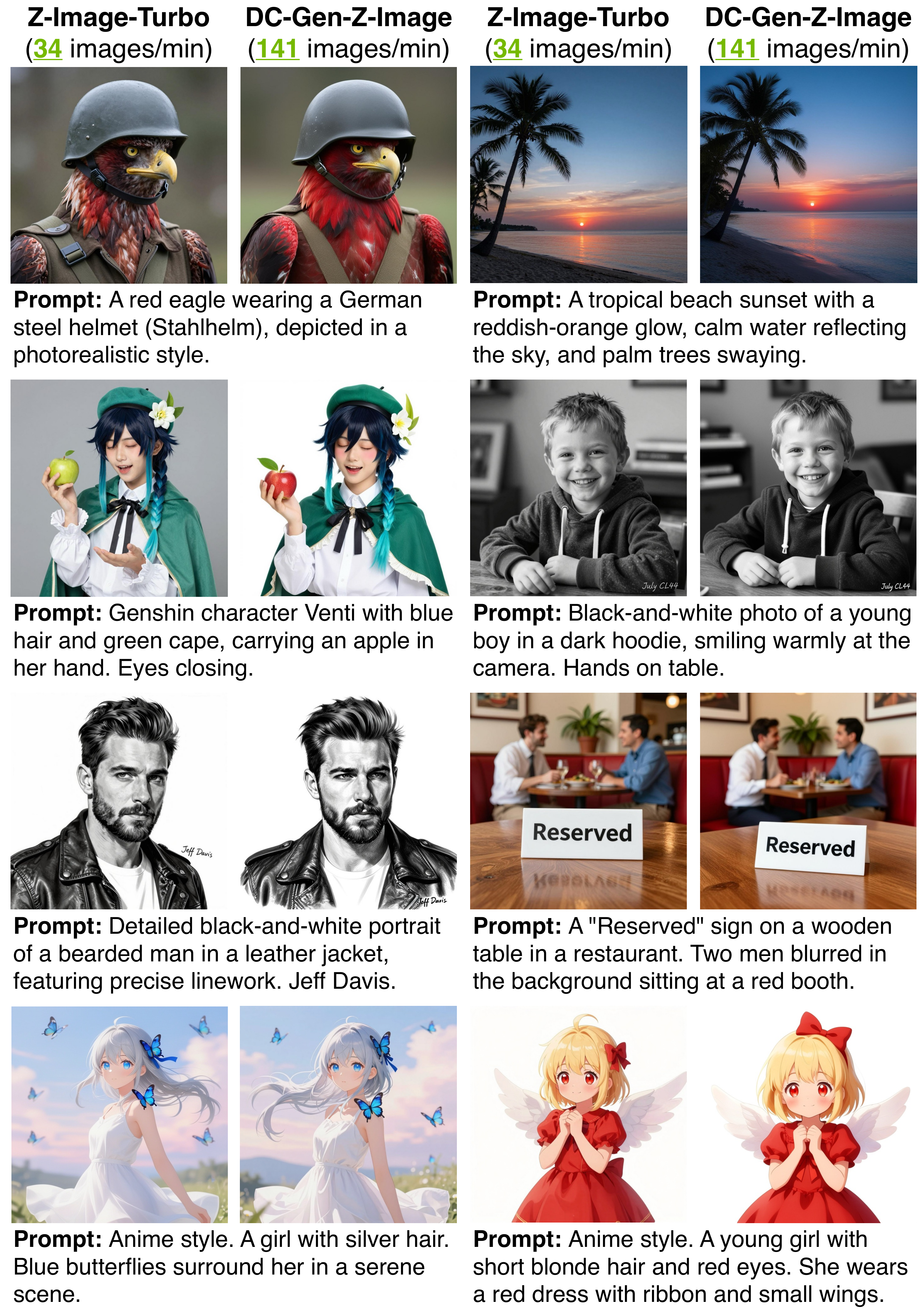}
    \caption{\textbf{1K resolution samples of Z-Image-Turbo and DC-Gen-Z-Image.}
    }
    \label{fig:appendix_zimage_sample1k}
\end{figure}

We demonstrate 1K image pairs generated by DC-Gen models and their corresponding pre-trained models in \figref{fig:appendix_flux_sample1k} and \figref{fig:appendix_zimage_sample1k}.

\begin{figure}[htbp]
    \centering
    \includegraphics[width=\linewidth]{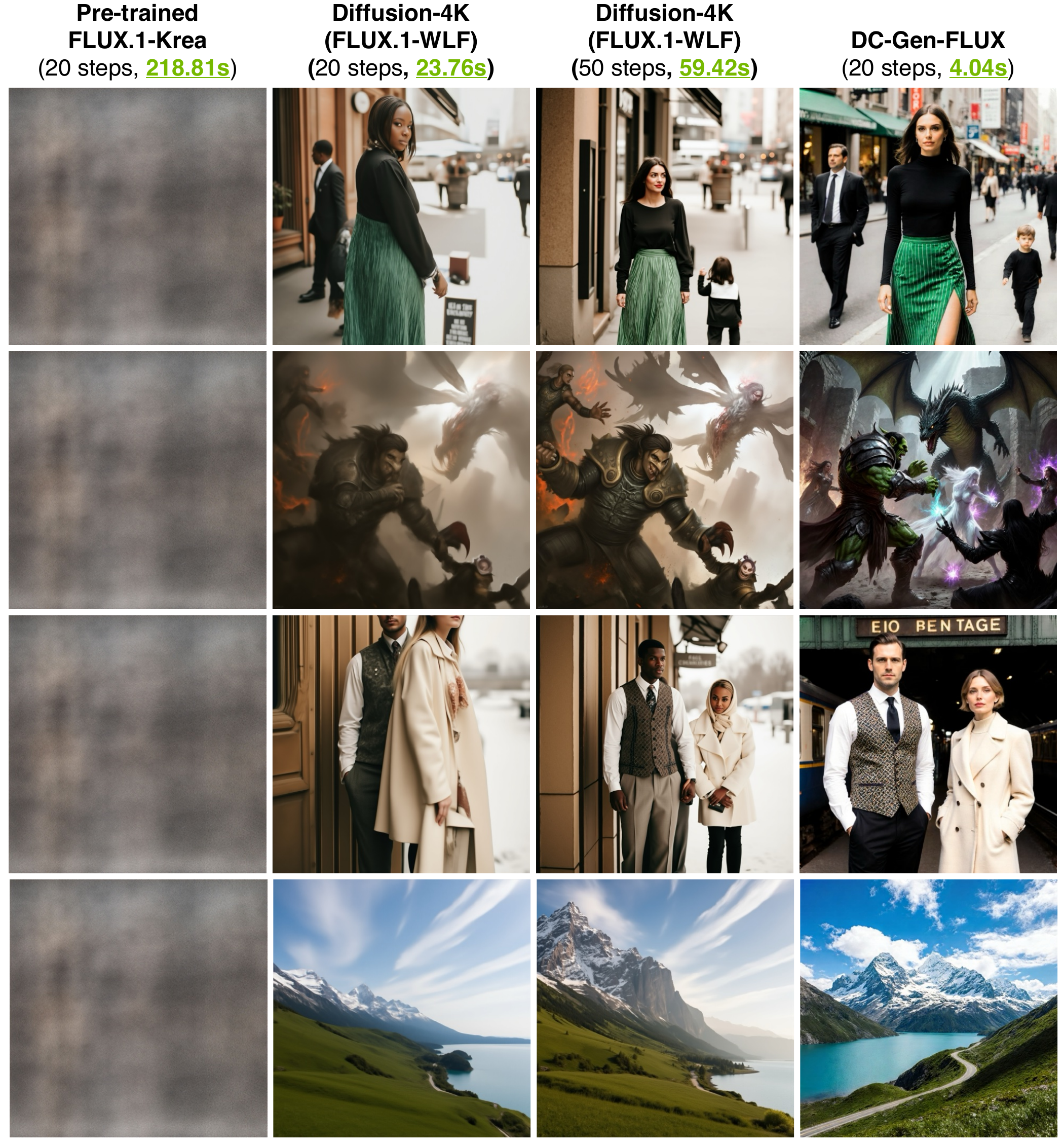}
    \caption{\textbf{Visual Results on the Diffusion-4K Benchmark.} }
    \label{fig:appendix_flux_sample4k}
\end{figure}

Additionally, we compare 4K image generation results in \figref{fig:appendix_flux_sample4k}. Pre-trained FLUX.1-Krea does not support native 4K image generation. Comparing with the state-of-the-art 4K image generation method Diffusion-4K~\cite{zhang2025diffusion4k}, DC-Gen achieves much better visual quality with significantly higher efficiency, demonstrating the potential of deeply compressed latent spaces.

\subsection{Text-to-Video and Image-to-Video Generation}

We compare the videos generated by DC-Gen-Wan2.1-T2V-14B and DC-Gen-Wan2.1-I2V-14B with their corresponding pre-trained Wan2.1-14B models on the webpage in the supplementary material.
\subsection{Image Editing}
\label{sec:appendix_image_edit_samples}
\definecolor{deeppink}{RGB}{255,105,180} 
\begin{figure}[htbp]
    \centering
    \includegraphics[width=\linewidth]{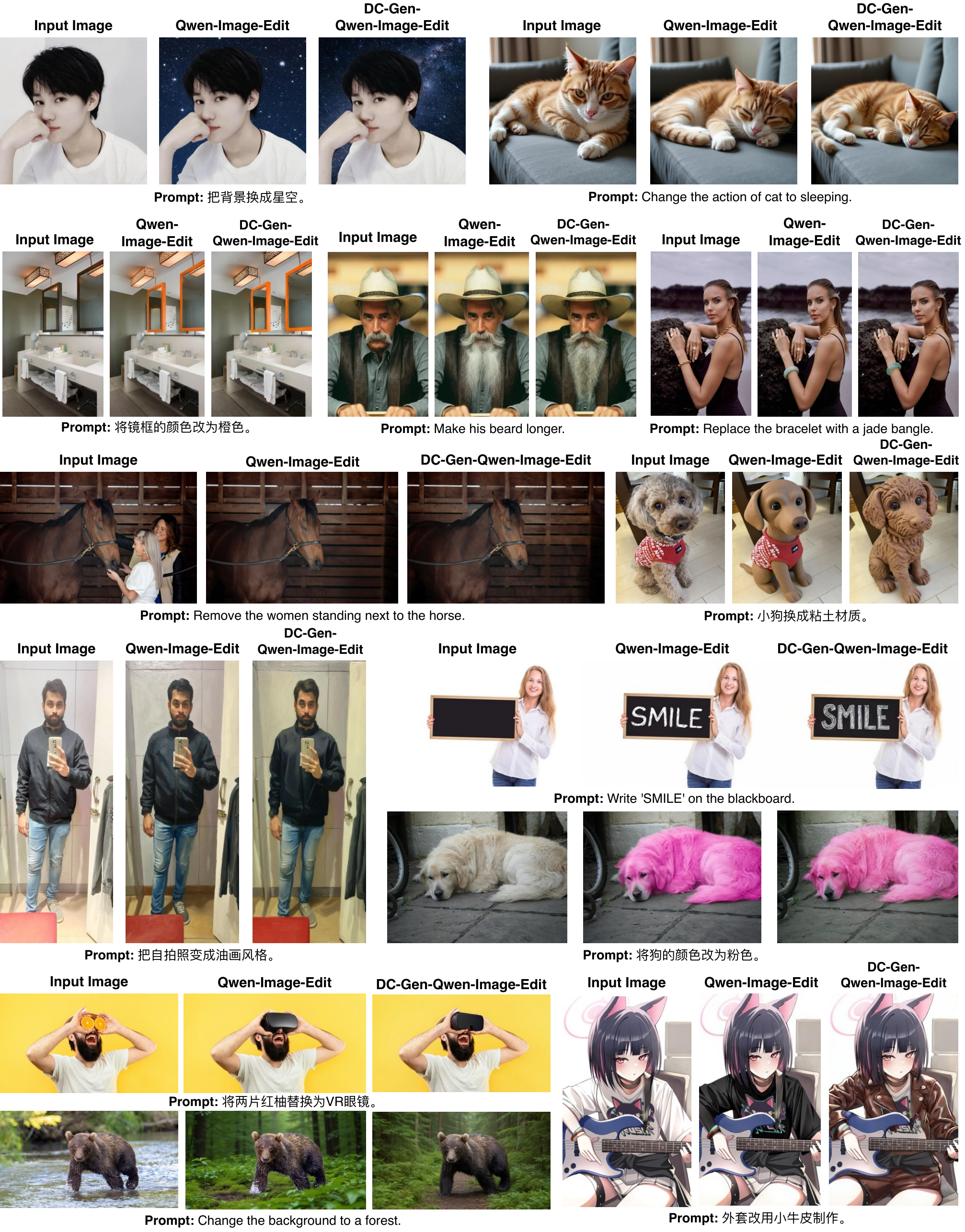}
    \caption{\textbf{Samples of Qwen-Image-Edit and DC-Gen-Qwen-Image-Edit on 1K Resolution Image Editing.} 
    }
    \label{fig:appendix_qwen_sample1k}
\end{figure}

We demonstrate additional image editing samples in \figref{fig:appendix_qwen_sample1k}. They are picked from our samples for human evaluation. DC-Gen-Qwen-Image-Edit supports image editing with different resolution ratios in both English and Chinese, which achieves comparable image quality with Qwen-Image-Edit.

\clearpage

\section{Human Evaluation}

\subsection{Text-to-Image Generation}
\label{sec:human_study_t2i}

\begin{figure}[htbp]
    \centering
    \includegraphics[width=\linewidth]{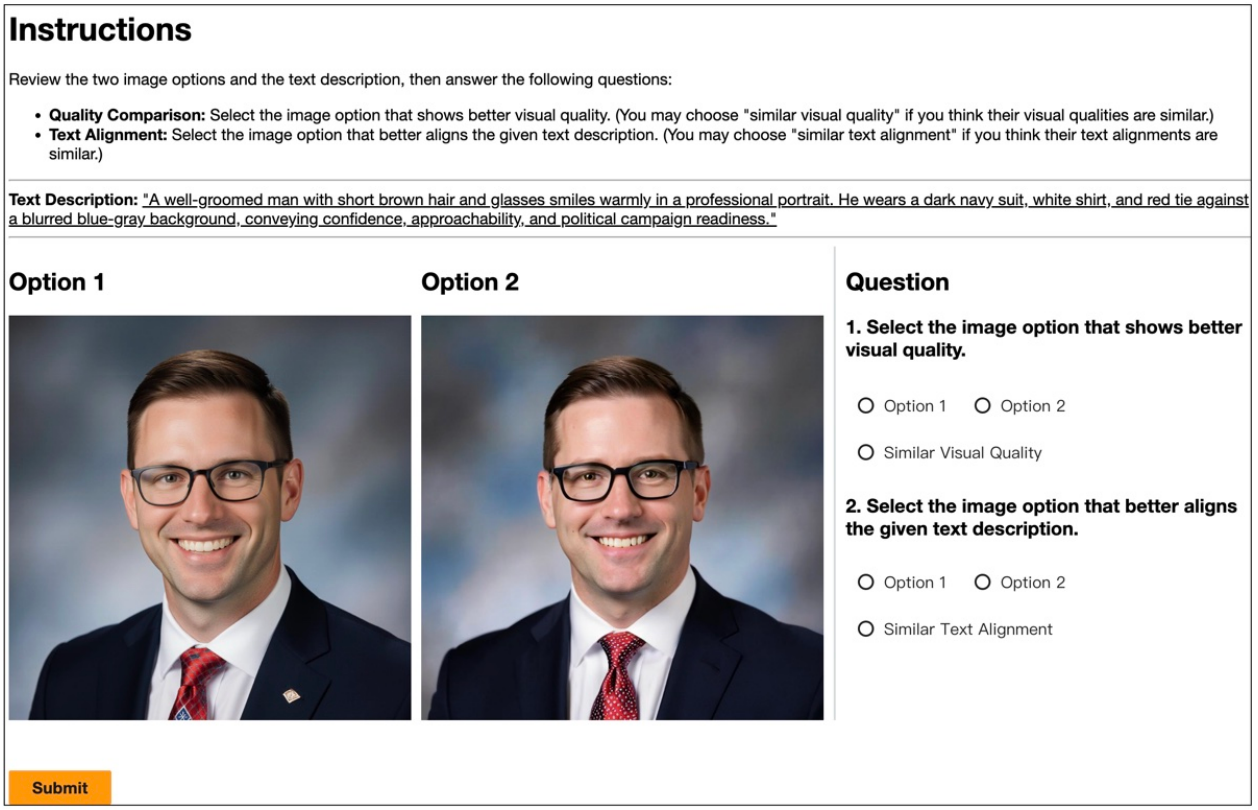}
    \vspace{-.2in}
    \caption{\textbf{The Human Evaluation Interface for Text-to-Image Generation.
    }}
    \label{fig:appendix_t2i_interface}
\end{figure}

In \secref{T2I Gen}, we've provided the human study results of DC-Gen-FLUX against the pre-trained FLUX.1-Krea. An example user interface is demonstrated in \figref{fig:appendix_t2i_interface}. Here, we further compare DC-Gen-Z-Image with the pre-trained Z-Image-Turbo in both visual quality and text alignment. Similarly, we selected 3 categories: general, anime, and real human, with 25 image pairs per category. Each pair was independently presented to 20 users. The results are shown in \tabref{tab:appendix_t2i_user_study}. DC-Gen-Z-Image achieves similar visual quality and text alignment to the pre-trained Z-Image-Turbo.

\begin{table}[htbp]
\vspace{-.05in}
\caption{\textbf{Human Evaluation Results for Z-Image.}}
\vspace{-5pt}
\small\centering\setlength{\tabcolsep}{2.5pt}
\resizebox{0.95\linewidth}{!}{
\begin{tabular}{l | c c c | c c c}
\toprule
& \multicolumn{3}{c|}{\textbf{Visual Quality}} & \multicolumn{3}{c}{\textbf{Text Alignment}} \\\cmidrule{2-7} & \textbf{Z-Image-Turbo} & & \textbf{DC-Gen-Z-Image}  & \textbf{Z-Image-Turbo} & & \textbf{DC-Gen-Z-Image} \\
\multirow{-3}{*}{\textbf{Domain}} & \textbf{is better}  & \multirow{-2}{*}{\textbf{Equal}} & \textbf{is better} & \textbf{is better} & \multirow{-2}{*}{\textbf{Equal}} & \textbf{is better} \\
\midrule
General & 44.2\% & 10.6\% & 45.2\% & 43.0\% & 14.0\% & 43.0\% \\
Person & 42.6\% & 13.2\% & 44.2\% & 45.6\% & 14.8\% & 39.6\% \\
Anime & 46.2\% & 12.0\% & 41.8\% & 43.0\% & 12.4\% & 44.6\% \\
\midrule
All & 44.3\% & 11.9\% & 43.8\% & 43.9\% & 13.7\% & 42.4\% \\
\bottomrule
\end{tabular}}
\label{tab:appendix_t2i_user_study}
\end{table}

\subsection{Text-to-Video and Image-to-Video Generation}
\label{sec:human_study_videogen}
\begin{figure}[htbp]
    \centering
    \includegraphics[width=\linewidth]{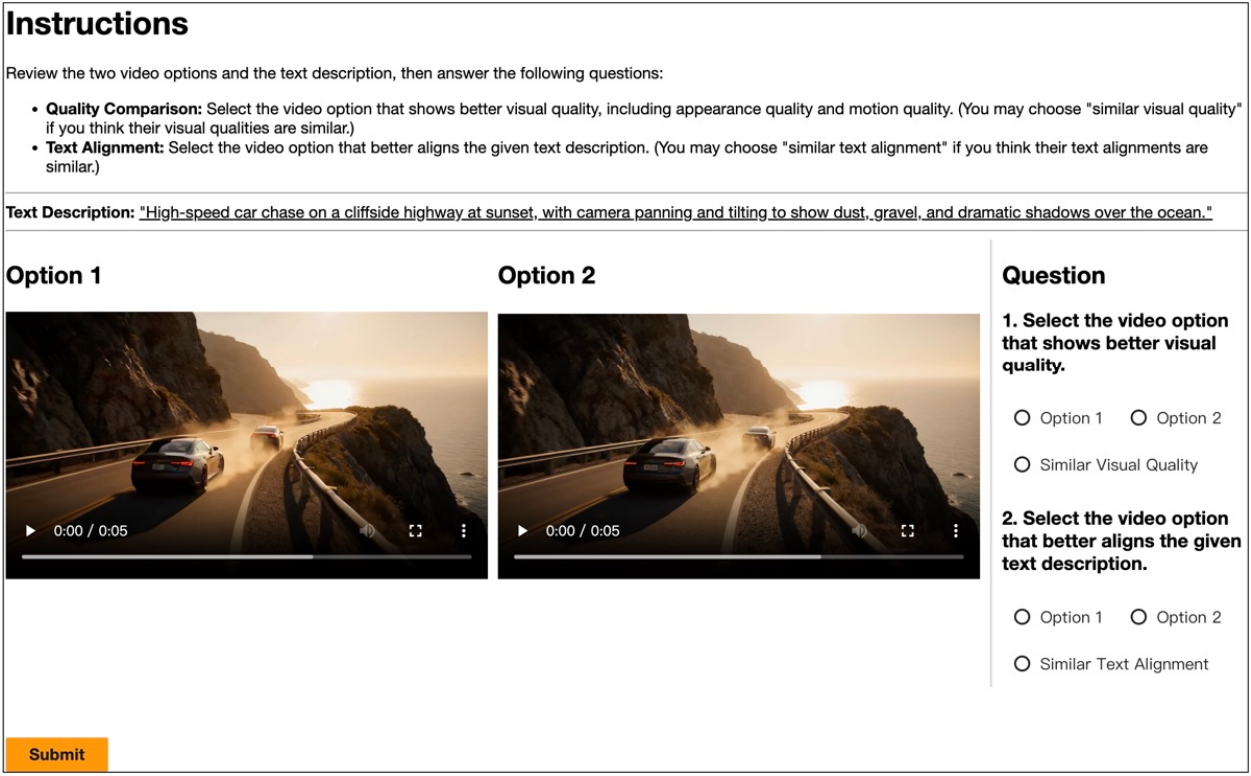}
    \vspace{-.2in}
    \caption{\textbf{The Human Evaluation Interface for Video Generation.
    }}
    \vspace{-.15in}
    \label{fig:appendix_videogen_interface}
\end{figure}

As shown in \figref{fig:appendix_videogen_interface}, we conducted a human study on both text-to-video and image-to-video tasks. We selected 20 pairs of 720p videos generated by DC-Gen-Wan2.1 and pre-trained Wan2.1 for both T2V and I2V. Each pair was independently assigned to 30 users. Users were asked to choose the video with higher visual quality or closer text alignment. The results are demonstrated in \tabref{tab:appendix_video_gen_user_study}, where DC-Gen achieves a slightly higher win rate. This is because the videos generated by DC-Gen models exhibit higher brightness and greater motions, which are preferred by average users. This is also illustrated in the video samples provided in the supplementary materials.
\begin{table}[htbp]
\small\centering\setlength{\tabcolsep}{1pt}
\caption{\textbf{Human Evaluation Results for Video Generation.}}
\vspace{-5pt}
\resizebox{0.95\linewidth}{!}{
\begin{tabular}{l|g g g | g g g}
\toprule
\multicolumn{7}{l}{\textbf{Human Evaluation Results on 720P Video Generation}} \\ 
\midrule
\rowcolor{white}  & \multicolumn{3}{c|}{\textbf{Visual Quality}} & \multicolumn{3}{c}{\textbf{Text Alignment}} \\\cmidrule{2-7}
\rowcolor{white}   & \textbf{Wan2.1-14B} & & \textbf{DC-Gen-Wan2.1-14B}  & \textbf{Wan2.1-14B} &  & \textbf{DC-Gen-Wan2.1-14B} \\
\rowcolor{white}  \multirow{-4}{*}{\textbf{Task}} & \textbf{is better} & \multirow{-2}{*}{\textbf{Equal}} & \textbf{is better}  & \textbf{is better}& \multirow{-2}{*}{\textbf{Equal}}  & \textbf{is better} \\\midrule
\rowcolor{white} Text-to-Video  & 34.8\% & 24.8\% & 40.4\%  & 33.7\% & 25.3\% & 41.0\% \\
\rowcolor{white} Image-to-Video & 35.0\% & 28.3\% & 36.7\% & 33.5\% & 28.5\% & 38.0\% \\
\bottomrule
\end{tabular}}
\vspace{-.05in}
\label{tab:appendix_video_gen_user_study}
\end{table}

\subsection{Image Editing}
\label{sec:human_study_image_edit}
\begin{table}[t]
\vspace{-.1in}
\caption{\textbf{Human Evaluation Results for Image Editing.}}
\vspace{-5pt}
\small\centering\setlength{\tabcolsep}{2.5pt}
\resizebox{0.95\linewidth}{!}{
\begin{tabular}{l | c c c | c c c}
\toprule
& \multicolumn{3}{c|}{\textbf{Visual Quality}} & \multicolumn{3}{c}{\textbf{Text Alignment}} \\\cmidrule{2-7}
 & \textbf{Qwen-Image-Edit} & & \textbf{DC-Gen-Qwen}  & \textbf{Qwen-Image-Edit} & & \textbf{DC-Gen-Qwen} \\
\multirow{-3}{*}{\textbf{Language}} & \textbf{is better}  & \multirow{-2}{*}{\textbf{Equal}} & \textbf{is better} & \textbf{is better}  & \multirow{-2}{*}{\textbf{Equal}} & \textbf{is better}\\
\midrule
English & 47.4\% & 10.2\% & 42.4\% & 46.0\% & 10.8\% & 43.2\% \\
Chinese & 42.2\% & 13.8\% & 44.0\% & 40.0\% & 13.4\% & 46.6\% \\
\midrule
All  & 44.8\% & 12.0\% & 43.2\%  & 43.0\% & 12.1\% & 44.9\%\\
\bottomrule
\end{tabular}}
\vspace{-.1in}
\label{tab:appendix_edit_user_study}
\end{table}

\begin{figure}[htbp]
    \centering
    \includegraphics[width=\linewidth]{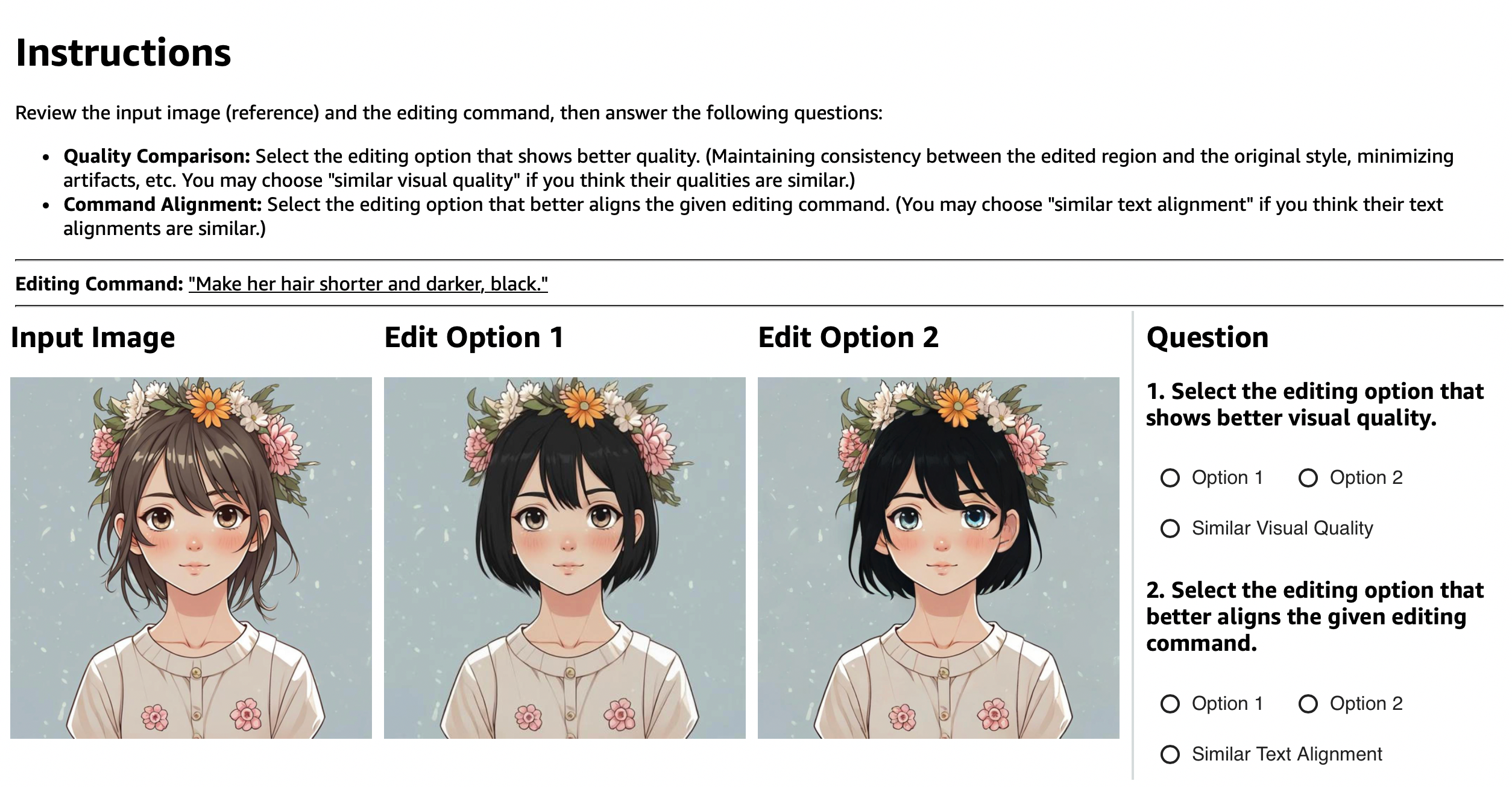}
    \vspace{-.25in}
    \caption{\textbf{The Human Evaluation Interface for Image Editing.
    }}
    \vspace{-.15in}
    \label{fig:appendix_editing_interface}
\end{figure}

Similar to text-to-image generation and video generation, we conducted a human study between DC-Gen-Qwen-Image-Edit and Qwen-Image-Edit (tuned on our training dataset) for 1K resolution image editing. We randomly selected 50 samples from our evaluation dataset GEdit~\cite{liu2025step1x-edit} (\secref{Image Editing}), with 25 in English and 25 in Chinese (We translated them into English during the human study). Each sample was presented to 20 different users. Users were required to answer two questions: which edited image has higher visual quality, and which edited image is more aligned with the given editing instruction. The results are demonstrated in \tabref{tab:appendix_edit_user_study}, where DC-Gen exhibits visual effects and text alignment that approach those of the pre-trained model for both English and Chinese prompts. We provide the sample user interface in \figref{fig:appendix_editing_interface}.

\clearpage
\section{Hyperparameters}
\label{sec:hyper_params}

\tabref{tab:hyper_params_t2i}, \tabref{tab:hyper_params_video}, and \tabref{tab:hyper_params_edit} demonstrate detailed training hyperparameters for text-to-image, video generation, and image editing tasks. For text-to-image generation and image editing, the standard resolution is 1K, while for text-to-video and image-to-video generation, the standard configuration is 720p, 80 frames. To reduce training costs, we perform the three-stage pipeline of DC-Gen at lower resolutions (512px for image generation and editing, and 480p for video generation). After convergence, we directly apply end-to-end LoRA fine-tuning at higher resolutions, a process we refer to as \textbf{resolution lifting}.

\begin{table}[htbp]
\caption{\textbf{Hyperparameters for the Training of DC-Gen-FLUX and DC-Gen-Z-Image.}}
\centering
\resizebox{0.9\linewidth}{!}{
\begin{tabular}{g | c | c | c }
\toprule
\rowcolor{white} \myPara{Training Stage} & \myPara{Hyper-parameter} & \myPara{FLUX.1-Krea~\cite{flux1kreadev2025}} & \myPara{Z-Image-Turbo~\cite{imageteam2025zimageefficientimagegeneration}}\\
\midrule
\rowcolor{white}  & learning rate & 1e-4 & 1e-4 \\
\rowcolor{white} & warmup steps & 0 & 0\\
\rowcolor{white} & batch size & 64 & 64 \\
\rowcolor{white} \multirow{-4}{*}{Patch Embedder} & training steps & 20k & 20k\\
\rowcolor{white} \multirow{-4}{*}{Alignment}& optimizer & AdamW, betas=[0.9, 0.999] & AdamW, betas=[0.9, 0.999]\\
\midrule
\rowcolor{white}  & learning rate  & 1e-4 & 1e-4\\ 
\rowcolor{white} & warmup steps  & 0 & 0\\
\rowcolor{white} \multirow{-2}{*}{Output Layer}& batch size  & 512 & 512 \\
\rowcolor{white} \multirow{-2}{*}{Alignment} & training steps  & 5k & 5k\\
\rowcolor{white} & optimizer & AdamW, betas=[0.9, 0.999] & AdamW, betas=[0.9, 0.999]\\
\midrule
\rowcolor{white}  & learning rate  & 1e-4 & 1e-4\\
\rowcolor{white} & warmup steps & 2k & 1k\\
\rowcolor{white}  & training steps  & 10k & 10k \\
\rowcolor{white} \multirow{-1}{*}{End-to-End} & batch size & 512 & 512 \\
\rowcolor{white} \multirow{-1}{*}{Fine-Tuning} & optimizer & AdamW, betas=[0.9, 0.999] & AdamW, betas=[0.9, 0.999]\\
\rowcolor{white} & weight decay & 1e-3 & 0\\
\rowcolor{white} & ema & 0.999 & 0.999\\
\rowcolor{white} & LoRA (rank, alpha) & (256, 256) & (256, 256) \\
\midrule
\rowcolor{white}  & learning rate  & 1e-4 & 1e-4\\
\rowcolor{white} & warmup steps & 0 & 0\\
\rowcolor{white}  & training steps  & 5k & 3k \\
\rowcolor{white} \multirow{-1}{*}{Resolution} & batch size & 256 & 256 \\
\rowcolor{white} \multirow{-1}{*}{Lifting} & optimizer & AdamW, betas=[0.9, 0.999] & AdamW, betas=[0.9, 0.999]\\
\rowcolor{white} & weight decay & 1e-3 & 0\\
\rowcolor{white} & ema & 0.999 & 0.999\\
\rowcolor{white} & LoRA (rank, alpha) & (256, 256) & (256, 256) \\
\bottomrule
\end{tabular}
}
\vspace{-5pt}
\label{tab:hyper_params_t2i}
\end{table}
\begin{table}[htbp]
\caption{\textbf{Hyperparameters for the Training of DC-Gen-Wan2.1.}}
\small\centering\setlength{\tabcolsep}{2pt}
\resizebox{0.78\linewidth}{!}{
\begin{tabular}{l | c | c}
\toprule
\rowcolor{white} Training Stage & Hyperparameter & Value \\
\midrule
\rowcolor{white}  & learning rate & 1e-4\\
\rowcolor{white} & warmup steps & 0 \\
\rowcolor{white} & batch size & 4 \\
\rowcolor{white} \multirow{-3}{*}{Patch Embedder Alignment} & training steps & 20k \\
\rowcolor{white} & optimizer & AdamW, betas=[0.9, 0.999] \\
\midrule
\rowcolor{white}  & learning rate & 1e-4 \\
\rowcolor{white} & warmup steps & 0 \\
\rowcolor{white} & batch size & 32 \\
\rowcolor{white} \multirow{-3}{*}{Output Layer Alignment} & training steps & 4k (Wan2.1-1.3B) / 3k (Wan2.1-14B)\\
\rowcolor{white} & optimizer & AdamW, betas=[0.9, 0.999] \\
\midrule
\rowcolor{white}  & learning rate & 5e-5 \\
\rowcolor{white} & warmup steps & 1k \\
\rowcolor{white} & training steps & 20k (Wan2.1-1.3B) / 6k (Wan2.1-14B) \\
\rowcolor{white} & batch size & 32 \\
\rowcolor{white} & optimizer & AdamW, betas=[0.9, 0.999] \\
\rowcolor{white} & weight decay & 1e-3 \\
\rowcolor{white} \multirow{-7}{*}{End-to-End Fine-Tuning} & LoRA (rank, alpha) & (256, 512) \\
\midrule
\rowcolor{white}  & learning rate & 5e-5 \\
\rowcolor{white} & warmup steps & 0 \\
\rowcolor{white} & training steps & 1k \\
\rowcolor{white} & batch size & 32 \\
\rowcolor{white} & optimizer & AdamW, betas=[0.9, 0.999] \\
\rowcolor{white} & weight decay & 1e-3 \\
\rowcolor{white} \multirow{-7}{*}{Resolution Lifting} & LoRA (rank, alpha) & (256, 512) \\
\bottomrule
\end{tabular}}
\label{tab:hyper_params_video}
\end{table}

\begin{table}[htbp]
\caption{\textbf{Training Hyperparameters for DC-Gen-Qwen-Image-Edit.}}
\small\centering\setlength{\tabcolsep}{2pt}
\resizebox{0.7\linewidth}{!}{
\begin{tabular}{l | c | c}
\toprule
\rowcolor{white} Training Stage & Hyperparameter & Value \\
\midrule
\rowcolor{white}  & learning rate & 1e-4\\
\rowcolor{white} & warmup steps & 0 \\
\rowcolor{white} & batch size & 64 \\
\rowcolor{white} \multirow{-3}{*}{Patch Embedder Alignment} & training steps & 50k \\
\rowcolor{white} & optimizer & AdamW, betas=[0.9, 0.999] \\
\midrule
\rowcolor{white}  & learning rate & 1e-4 \\
\rowcolor{white} & warmup steps & 0 \\
\rowcolor{white} & batch size & 256 \\
\rowcolor{white} \multirow{-3}{*}{Output Layer Alignment} & training steps & 5k \\
\rowcolor{white} & optimizer & AdamW, betas=[0.9, 0.999] \\
\midrule
\rowcolor{white}  & learning rate & 1e-4 \\
\rowcolor{white} & warmup steps & 1k \\
\rowcolor{white} & training steps & 10k \\
\rowcolor{white} & batch size & 256 \\
\rowcolor{white} & optimizer & AdamW, betas=[0.9, 0.999] \\
\rowcolor{white} & weight decay & 1e-3 \\
\rowcolor{white} \multirow{-7}{*}{End-to-End Fine-Tuning} & LoRA (rank, alpha) & (256, 256) \\
\midrule
\rowcolor{white}  & learning rate & 1e-4 \\
\rowcolor{white} & warmup steps & 0 \\
\rowcolor{white} & training steps & 5k \\
\rowcolor{white} & batch size & 256 \\
\rowcolor{white} & optimizer & AdamW, betas=[0.9, 0.999] \\
\rowcolor{white} & weight decay & 1e-3 \\
\rowcolor{white} \multirow{-7}{*}{Resolution Lifting} & LoRA (rank, alpha) & (256, 256) \\
\bottomrule
\end{tabular}}
\label{tab:hyper_params_edit}
\end{table}

\clearpage

\section{Detailed Training and Inference Cost}
\subsection{Training Cost}
\label{sec:appendix_training_cost}
\begin{figure}[htbp]
    \centering
    \includegraphics[width=0.77\linewidth]{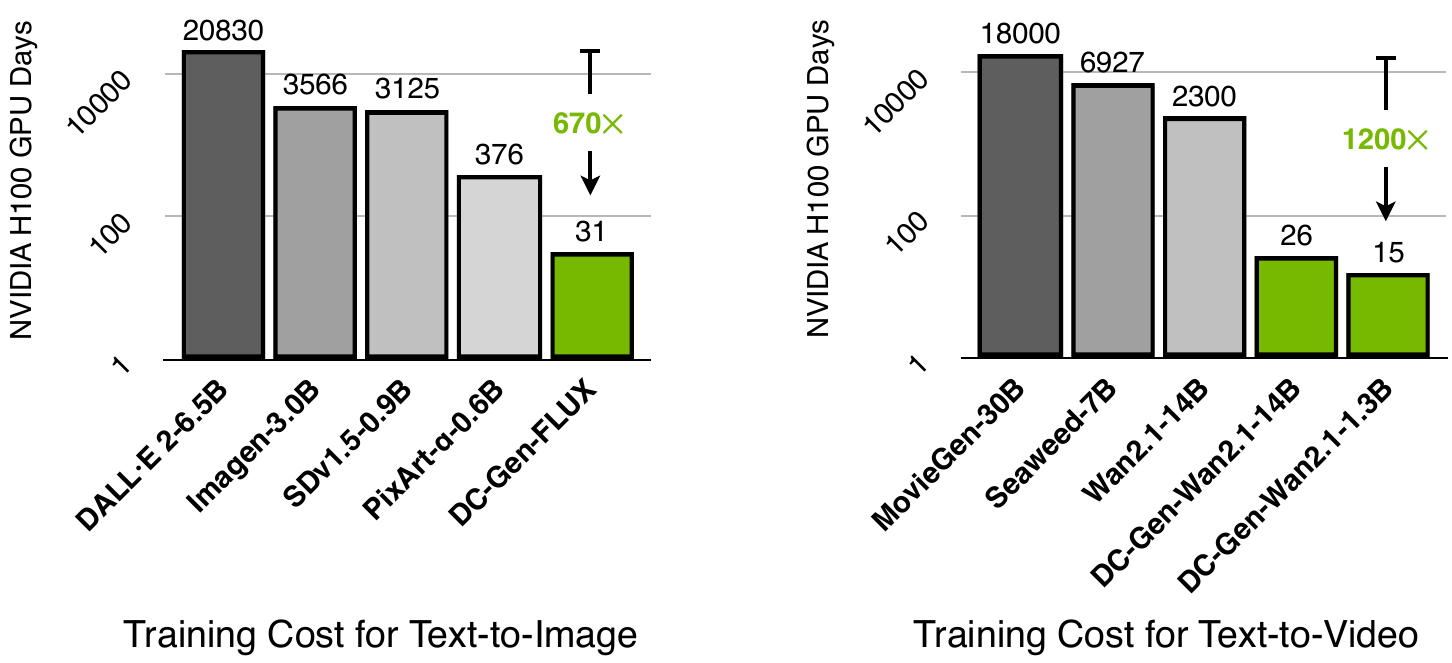}
    \vspace{-.1in}
    \caption{
         \textbf{Training cost of DC-Gen, measured in H100 GPU hours.}}
    \label{fig:appendix_training_cost}
    \vspace{-.1in}
\end{figure}

\begin{table}[htbp]
\centering
\caption{\textbf{Training cost for different DC-Gen Models.} The cost is measured in NVIDIA H100 GPU Days. $^\dag$Resolution lifting is defined in \secref{sec:hyper_params}.}
\label{tab:training_cost}
\resizebox{\linewidth}{!}{
\begin{tabular}{g|g|g| g g g g}
\toprule
\rowcolor{white} & & & \textbf{Patch Embedder} & \textbf{Output Layer} & \textbf{End-to-End} & \textbf{Resolution} \\
\rowcolor{white} \multirow{-2}{*}{\textbf{Base Model}} & \multirow{-2}{*}{\textbf{Resolution}} & \multirow{-2}{*}{\textbf{Total}} & \textbf{Alignment} & \textbf{Alignment} & \textbf{Fine-Tuning} & \textbf{Lifting}$^\dag$ \\
\midrule
\rowcolor{white} DC-Gen-FLUX & 1K & 31.25  & 0.14 & 5.93 & 14.39 & 10.80  \\
\rowcolor{white} DC-Gen-Z-Image & 1K & 26.90  & 0.08 & 6.46 & 14.79 & 5.57 \\
\midrule
\rowcolor{white} DC-Gen-Wan2.1-T2V-1.3B  & 720p & 14.90  & 0.05 & 1.78 & 11.13 & 1.94 \\
\rowcolor{white} DC-Gen-Wan2.1-T2V-14B & 720p  & 25.59 & 0.08 & 6.67 & 14.68 & 4.15 \\
\rowcolor{white} DC-Gen-Wan2.1-I2V-14B & 720p  & 29.77  & 0.10 & 8.33 & 16.67 & 4.67 \\
\midrule
\rowcolor{white} DC-Gen-Qwen-Image-Edit & 1K & 30.85 & 0.44 & 4.63 & 13.96 & 11.82 \\
\bottomrule
\end{tabular}}
\label{tab:appendix_training_cost}
\end{table}

We measured the training cost of DC-Gen in NVIDIA H100 GPU Days, which equals the number of GPUs multiplied by the actual training time. As demonstrated in \tabref{tab:appendix_training_cost} and \figref{fig:appendix_training_cost}, the post-training process of all DC-Gen models can be accomplished within 40 H100 GPU days, which is negligible considering the high cost of pre-training a model from scratch.

Furthermore, as shown in \tabref{tab:appendix_training_cost}, embedding alignment (consisting of the alignment of the patch embedder and output layer) accounts for only a small proportion of the total training cost, indicating that DC-Gen significantly improves the stability, quality, and convergence speed of the end-to-end fine-tuning process in autoencoder adaptation at a minimal additional cost.

\subsection{Inference Speedup}

\begin{figure}[htbp]
    \centering
    \includegraphics[width=\linewidth]{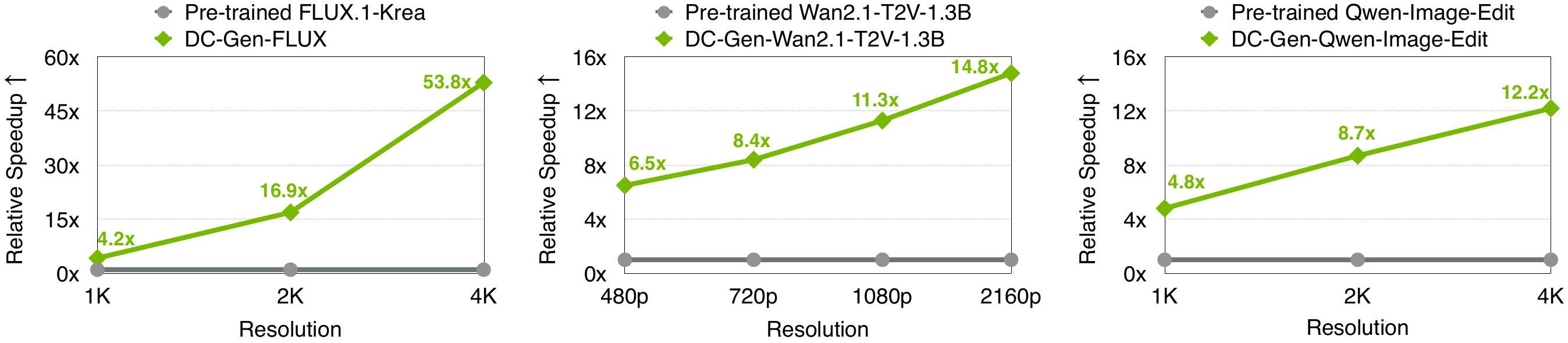}
    \vspace{-.15in}
    \caption{
         \textbf{Inference speedup of DC-Gen at different resolutions.}}
    \label{fig:appendix_inference_speedup}
    \vspace{-.05in}
\end{figure}

In \figref{fig:appendix_inference_speedup}, we demonstrate the relative inference latency speedup of DC-Gen across different resolutions. Since the number of text tokens remains constant regardless of the compression ratios, the proportion of visual tokens grows as the resolution increases, leading to a higher relative speedup for DC-Gen models.

In this paper, the latency is simplified as the cost within the transformer backbone. We further compare the latency of transformer denoising and autoencoder decoding in \tabref{tab:appendix_latency_breakup}. For the high-resolution generation that DC-Gen targets, the latter is nearly negligible compared to the former.
\begin{table}[h]
\caption{\textbf{Latency breakup for generating a single image on an H100 GPU.} $^\dag$We enable autoencoder tiling to prevent out-of-memory error.}
\centering
\resizebox{0.9\linewidth}{!}{
\begin{tabular}{l | c | c c | c c}
\toprule
\rowcolor{white} & & \multicolumn{2}{c|}{\myPara{Pre-trained Model}} & \multicolumn{2}{c}{\myPara{DC-Gen-Model}} \\
\rowcolor{white} \multirow{-2}{*}{\myPara{Backbone}} & \multirow{-2}{*}{\myPara{Resolution}} & Backbone & Autoencoder & Backbone  & Autoencoder \\
\midrule
& 1K & 4.06s & 0.15s & 0.88s & 0.29s \\
& 2K & 14.90s & 0.38s$^\dag$ & 0.88s & 0.55s\\
\multirow{-3}{*}{\textbf{FLUX.1-Krea}~\cite{flux1kreadev2025}} & 4K & 218.81s & 1.36s$^\dag$ & 4.04s & 0.95s \\
\midrule
& 720p & 5.76min & 0.05min & 0.70min & 0.12min \\
& 1080p & 25.40min & 0.22min & 3.18min & 0.30min \\
\multirow{-3}{*}{\textbf{Wan2.1-T2V-1.3B}~\cite{wan2025}} & 2160p & 230.44min & 0.94min$^\dag$ & 25.01min & 0.71min$^\dag$ \\
\midrule
& 1K & 0.96min & 0.00min & 0.20min & 0.01min \\
& 2K & 8.38min & 0.02min$^\dag$ & 0.96min & 0.02min \\
\multirow{-3}{*}{\textbf{Qwen-Image-Edit}~\cite{wu2025qwenimagetechnicalreport}} & 4K & 102.19min & 0.09min$^\dag$ & 8.38min & 0.06min$^\dag$ \\
\bottomrule
\end{tabular}
}
\vspace{-10pt}
\label{tab:appendix_latency_breakup}
\end{table}

\section{Combine DC-Gen with Orthogonal Techniques}
\label{sec:orthogonal}
There are two categories of approaches for accelerating latent diffusion models. The first one, typically represented by our proposed DC-Gen, reduces token counts through utilizing deeply compressed latent space, while the other directly accelerates the inference process itself. Representative methods of the latter include quantization, feature caching, few-step distillation, etc. These two categories are orthogonal to each other, and DC-Gen can be simply and effectively combined with methods such as SVDQuant (Quantization)~\cite{li2024svdquant}, TaylorSeer (TaylorSeer)~\cite{TaylorSeer2025}, and AnyFlow (4-step Distillation)~\cite{gu2026anyflow}, as shown in Table~\ref{tab:appendix_orthogonal_methods}. We conducted experiments with FLUX.1-Krea, 1024px.

\begin{table}[h]
\caption{\textbf{Inference Speed and CLIP Scores for Different Acceleration Methods.} }
\centering
\resizebox{0.9\linewidth}{!}{
\begin{tabular}{l | c | c | c | c }
\toprule
\rowcolor{white} \myPara{Backbone} & \myPara{Acceleration Method} & Latency (s) $\downarrow$ & Speedup $\uparrow$ & CLIP Score $\uparrow$ \\
\midrule
Pre-trained FLUX.1-Krea & - & 4.06 & 1.00$\times$ & 27.93 \\
DC-Gen-FLUX & - & 0.88 & 4.62$\times$ & 28.02\\
\midrule
& SVDQuant~\cite{li2024svdquant} & 2.37 & 1.71$\times$ & 27.78 \\
& TaylorSeer~\cite{TaylorSeer2025} & 1.47 & 2.76$\times$ & 27.55 \\
\multirow{-3}{*}{Pre-trained FLUX.1-Krea} & AnyFlow~\cite{gu2026anyflow} & 0.81 & 5.01$\times$ & 27.99 \\
\midrule
& SVDQuant~\cite{li2024svdquant} & 0.61 & 6.62$\times$ & 28.00 \\
& TaylorSeer~\cite{TaylorSeer2025} & 0.35 & 11.67$\times$ & 27.54 \\
\multirow{-3}{*}{DC-Gen-FLUX} & AnyFlow~\cite{gu2026anyflow} & \textbf{0.17} & \textbf{23.24$\times$} & \textbf{28.10}\\
\bottomrule
\end{tabular}
}
\vspace{-10pt}
\label{tab:appendix_orthogonal_methods}
\end{table}

\section{Deep Compression Video Autoencoder (DC-AE-V)}
\label{sec:appendix_dc_ae_v}

Existing video autoencoders can be categorized into two groups based on their temporal modeling designs: causal and non-causal.

\begin{figure}[htbp]
    \centering
    \includegraphics[width=\linewidth]{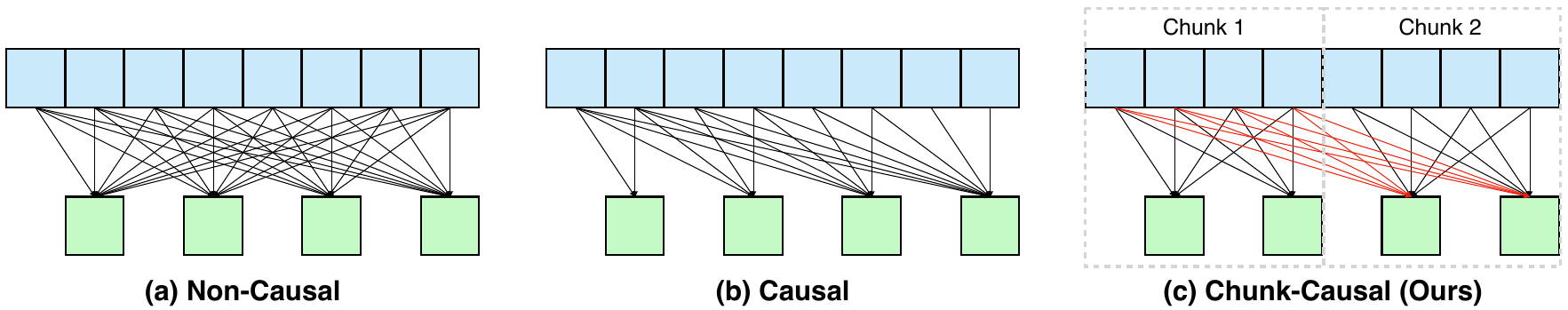}
    \caption{
        \textbf{Illustration of Chunk-Causal Temporal Modeling in DC-AE-V.} 
        Our chunk-causal temporal modeling preserves causal information flow across chunks while enabling bidirectional flow within each chunk. This design improves reconstruction quality over non-causal temporal modeling, while maintaining generalization to longer videos at inference time.
    }
    \vspace{-.1in}
    \label{fig:dc-ae-v}
\end{figure}

\begin{wrapfigure}{r}{0.3\textwidth}
\vspace{-30pt}
\begin{center}
    \includegraphics[width=0.3\textwidth]{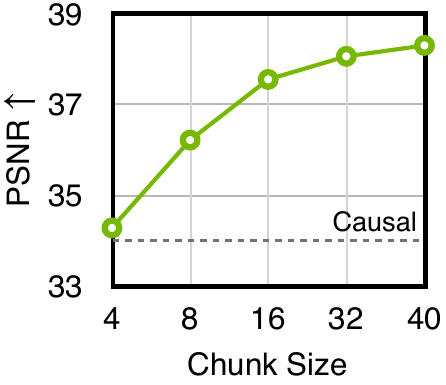}
\end{center}
\vspace{-15pt}
\caption{\textbf{Ablation Study on Chunk Size.}}
\vspace{-15pt}
\label{fig:ablation_chunk_size}
\end{wrapfigure}

\begin{figure}[t]
    \centering
    \includegraphics[width=0.9\linewidth]{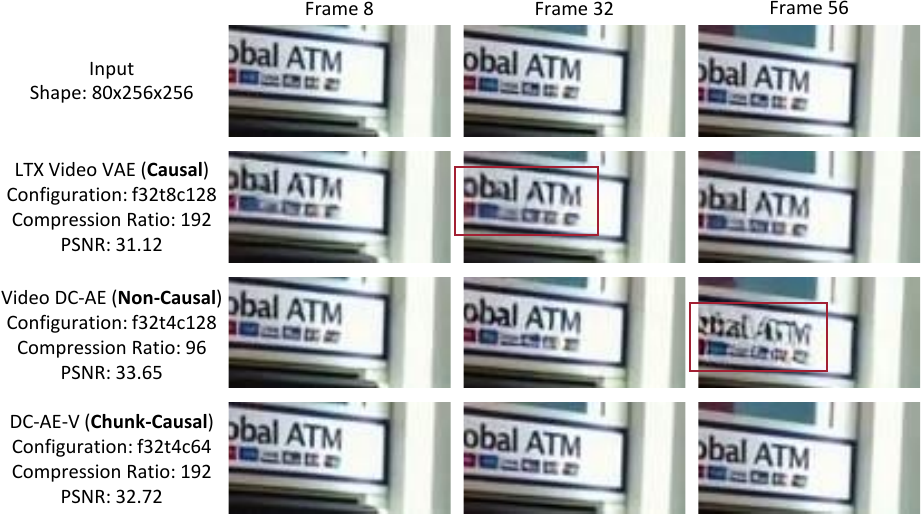}
    \vspace{-5pt}
    \caption{
        \textbf{Video Autoencoder Reconstruction Visualization.} 
        Under deep compression settings, causal video autoencoders suffer from low reconstruction quality. In contrast, non-causal video autoencoders achieve better reconstruction quality but generalize poorly to longer videos.
    }
    \vspace{-5pt}
    \label{fig:ae_visualization}
\end{figure}

\begin{itemize}[leftmargin=*]
\item \textbf{Causal.} In causal video autoencoders, information flows only from earlier frames to later frames (Figure~\ref{fig:dc-ae-v}b). This design naturally supports longer videos during inference, since the encoding and decoding of later frames do not affect earlier ones. However, because each frame can only leverage redundancy from preceding frames, reconstruction accuracy is limited under deep compression settings, as illustrated in Figure~\ref{fig:ae_visualization}. 

Building on the causal design, IV-VAE~\cite{wu2025improved} notes that, since every $t$ input frames are compressed into a single latent frame, enforcing causality within each group of $t$ frames is unnecessary. To address this, IV-VAE introduces a grouped causal convolution with group size $t$ to improve reconstruction performance. However, the group size is strictly tied to the temporal compression ratio, and as shown in Figure~\ref{fig:ablation_chunk_size}, it provides only limited improvements in reconstruction quality over the standard causal design under deep compression settings.
\item \textbf{Non-Causal.} Non-causal autoencoders allow bidirectional information flow between frames (Figure~\ref{fig:dc-ae-v}a). This enables each frame to leverage redundancy from both past and future frames, yielding better reconstruction quality under deep compression settings. However, because earlier frames depend on later ones, generalization to longer videos becomes challenging. Techniques such as temporal tiling and blending \cite{peng2025open} can partially alleviate this issue but often introduce artifacts, including temporal flickering and boundary blurring, as shown in Figure~\ref{fig:ae_visualization}.
\end{itemize}

We introduce a new temporal modeling design, \textbf{chunk-causal}, to overcome these limitations (Figure~\ref{fig:dc-ae-v}c). The key idea is to divide the input video into fixed-size chunks, where the chunk size is treated as an independent hyperparameter. Within each chunk, we apply bidirectional temporal modeling to fully exploit redundancy across frames. Across chunks, however, we enforce causal flow so that the model can effectively generalize to longer videos at inference time. Figure~\ref{fig:ablation_chunk_size} presents the ablation study on chunk size. We observe that increasing the chunk size consistently improves reconstruction quality. In our final design, we adopt a chunk size of 40, as the benefits plateau beyond this point while training costs continue to rise.

\begin{table}[h!]

\caption{
\textbf{Video Autoencoder Reconstruction Results.} 
$^\dag$Video DC-AE achieves higher PSNR with tiling and blending, but still exhibits poor generalization when applied to longer videos at inference time, as shown in Figure~\ref{fig:ae_visualization}. 
}
\vspace{-.05in}
\small\centering\setlength{\tabcolsep}{1pt}
\resizebox{\linewidth}{!}{
\begin{tabular}{f | g | g | g g g | g g g g}
\toprule
\rowcolor{white} &  & Compress. & \multicolumn{3}{c|}{Panda70m} & \multicolumn{4}{c}{UCF101} \\
\rowcolor{white} \multirow{-2}{*}{Video Autoencoder} & \multirow{-2}{*}{Config} & Ratio & PSNR$\uparrow$ & SSIM$\uparrow$ & LPIPS$\downarrow$ & PSNR$\uparrow$ & SSIM$\uparrow$ & LPIPS$\downarrow$ & FVD$\downarrow$ \\
\midrule
\rowcolor{white} VideoVAEPlus \cite{xing2024large}                    & f8t4c16   &  48 & 36.88 & 0.968 & 0.009 & 35.79 & 0.959 & 0.016 &  2.11 \\
\rowcolor{white} CogVideoX VAE \cite{yang2024cogvideox}               & f8t4c16   &  48 & 35.54 & 0.961 & 0.021 & 34.53 & 0.949 & 0.034 &  8.32 \\
\rowcolor{white} HunyuanVideo VAE \cite{kong2024hunyuanvideo}         & f8t4c16   &  48 & 35.46 & 0.960 & 0.015 & 34.40 & 0.950 & 0.024 &  3.80 \\
\rowcolor{white} IV VAE \cite{wu2025improved}                         & f8t4c16   &  48 & 35.32 & 0.959 & 0.017 & 34.84 & 0.955 & 0.025 &  3.71 \\
\rowcolor{white} Wan 2.1 VAE \cite{wan2025}                           & f8t4c16   &  48 & 34.15 & 0.952 & 0.017 & 33.81 & 0.943 & 0.024 &  3.71 \\
\rowcolor{white} Wan 2.2 VAE \cite{wan2025}                           & f16t4c48  &  64 & 35.12 & 0.958 & 0.013 & 34.27 & 0.948 & 0.022 &  4.02 \\
\rowcolor{white} StepVideo VAE \cite{ma2025step}                      & f16t8c64  &  96 & 32.17 & 0.930 & 0.043 & 32.17 & 0.930 & 0.043 &  8.23 \\
\midrule
\rowcolor{white} Video DC-AE$^{\dag}_{\scriptsize{\text{tiling \& blending}}}$ \cite{peng2025open} & f32t4c128 &  96 & 34.10 & 0.952 & 0.023 & 33.65 & 0.945 & 0.034 & 14.22 \\
\rowcolor{white} Video DC-AE \cite{peng2025open}                      & f32t4c128 &  96 & 31.73 & 0.915 & 0.040 & 31.52 & 0.914 & 0.047 & 26.30 \\
\rowcolor{white} LTX Video VAE \cite{hacohen2024ltx}                  & f32t8c128 & 192 & 32.41 & 0.928 & 0.039 & 31.12 & 0.910 & 0.059 & 70.92 \\
\midrule
                                                                           & f32t4c256 &  48 & 39.54 & 0.979 & 0.008 & 37.13 & 0.967 & 0.018 &  2.03 \\
                                                                           & f32t4c128 &  96 & 37.36 & 0.968 & 0.013 & 34.83 & 0.951 & 0.027 &  5.49 \\
                                                                           & f32t4c64  & 192 & 35.03 & 0.953 & 0.020 & 32.72 & 0.931 & 0.036 & 13.04 \\
                                                                           & f32t4c32  & 384 & 33.07 & 0.933 & 0.027 & 30.83 & 0.909 & 0.046 & 29.11 \\
\multirow{-5}{*}{DC-AE-V}                                                  & f64t4c128 & 384 & 32.83 & 0.932 & 0.030 & 30.65 & 0.907 & 0.049 & 32.16 \\
\bottomrule
\end{tabular}}
\vspace{-5pt}
\label{tab:video_ae_reconstruction}
\end{table}

\begin{table}[t]
\caption{
\textbf{Comparison of Video Generation Results across Different Autoencoders.} 
We adopt the same training setup for all models to ensure apples-to-apples comparisons, i.e., using DC-Gen to adapt the pretrained Wan-T2V-1.3B to different autoencoders.
}
\vspace{-.04in}
\small\centering\setlength{\tabcolsep}{4pt}
\resizebox{0.9\linewidth}{!}{
\begin{tabular}{l | g | g | g | g | g g}
\toprule
\multicolumn{7}{l}{\textbf{Text-to-Video Generation Results on VBench 480$\times$832}} \\
\midrule
\rowcolor{white} Text-to-Video & & Patch & Latency & \multicolumn{3}{c}{Score $\uparrow$} \\
\cmidrule{5-7}
\rowcolor{white} Diffusion Model & \multirow{-2}{*}{Video Autoencoder} & Size & (s) $\downarrow$ & Overall & Quality & Semantic \\
\midrule
\cellcolor{white} & \cellcolor{shadecolor} Wan2.1-VAE-f8t4c16~\cite{wan2025} & \cellcolor{shadecolor} 2 & \cellcolor{shadecolor} 89.30 & \cellcolor{shadecolor} 83.32 & \cellcolor{shadecolor} 85.01 & \cellcolor{shadecolor} 76.57 \\
\cmidrule{2-7}
\rowcolor{white} & LTX-Video-f32t8c128~\cite{hacohen2024ltx} & 1 & 6.98 & 82.21 & 83.97 & 75.16  \\
\rowcolor{white} & OpenSora2-f32t4c128~\cite{peng2025open} & 1 & 14.57 & 81.26 & 83.49 & 72.33 \\
\rowcolor{white} & Wan-2.2-VAE-f16t4c48~\cite{wan2025} & 2 & 14.59 & 80.38 & 83.20 & 69.08 \\
\cmidrule{2-7}
 \multirow{-6}{*}{Wan2.1-1.3B~\cite{wan2025}} & DC-AE-V-f32t4c32 & 1 & 14.55 & 83.48 & 85.08 & 77.12 \\
\bottomrule
\end{tabular}}
\label{tab:ablation_video_vae}
\end{table}

\myPara{Video Reconstruction Results.} 
We summarize the comparison between DC-AE-V and prior state-of-the-art video autoencoders in Table~\ref{tab:video_ae_reconstruction}. 
We use \textbf{f}x\textbf{t}y\textbf{c}z to denote the configuration of a video autoencoder. For example, \textbf{f8t4c16} represents a video autoencoder that compresses an input video of size $3 \times T \times H \times W$ into a latent of size $16 \times \frac{T}{4} \times \frac{H}{8} \times \frac{W}{8}$. 
The compression ratio is defined as
\begin{equation}
    \text{Compression Ratio} = \frac{3 T H W}{c \cdot \frac{T}{t} \cdot \frac{H}{f} \cdot \frac{W}{f}} = \frac{3 f^2 t}{c}.
\end{equation}
Compared with causal video autoencoders such as LTX Video VAE~\cite{hacohen2024ltx}, DC-AE-V achieves higher reconstruction accuracy at the same compression ratio, as well as higher compression ratios for a given accuracy target. Compared with non-causal video autoencoders such as Video DC-AE~\cite{peng2025open}, DC-AE-V delivers better reconstruction quality under the same compression ratio while also generalizing better to longer videos (Figure~\ref{fig:ae_visualization}).

\myPara{Video Generation Results.} 
In addition to reconstruction performance, we also evaluate DC-AE-V against prior autoencoders on video generation. Table~\ref{tab:ablation_video_vae} reports ablation results on Wan2.1-1.3B~\cite{wan2025}, showing that DC-AE-V achieves the best video generation performance. Compared with the base model, DC-AE-V provides a 6.2$\times$ speedup while attaining slightly higher VBench scores.

\myPara{Model Architecture.}
\begin{figure}[t]
    \centering
    \includegraphics[width=\linewidth]{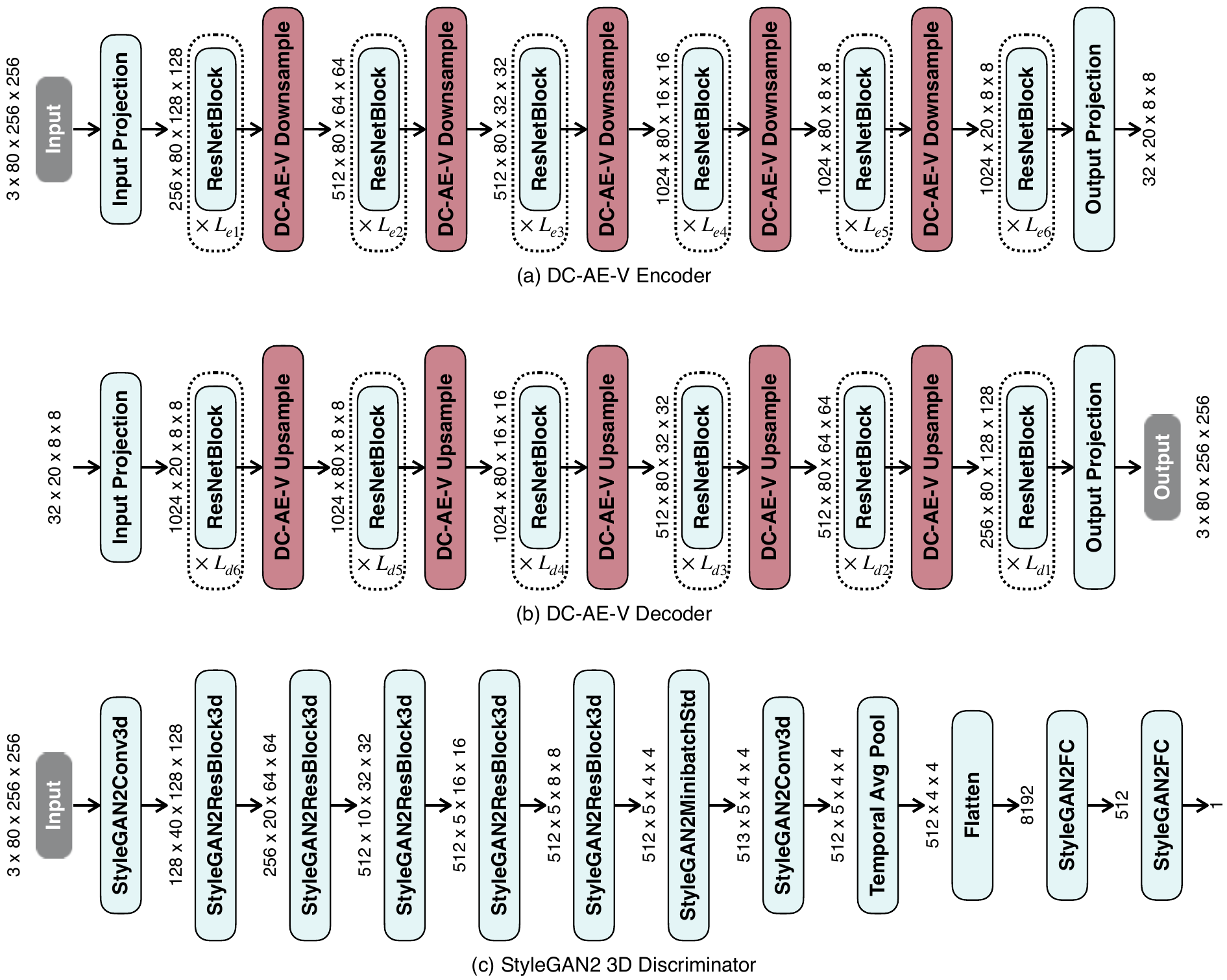}
    \caption{
        \textbf{Model Architecture of DC-AE-V.}
    }
    \label{fig:ae_architecture}
\end{figure}

Figure~\ref{fig:ae_architecture} presents the detailed model architecture of an f32t4c32 DC-AE-V. Both the encoder and decoder are composed of six stages, each built from 3D ResNet~\cite{he2016deep} blocks. The first five stages perform only spatial downsampling and upsampling, while the final stage handles temporal downsampling and upsampling. Following DC-AE~\cite{chen2024deep}, we incorporate Residual Autoencoding to facilitate optimization during downsampling and upsampling. For adversarial training, we extend the StyleGAN2 discriminator~\cite{karras2020analyzing} to process video inputs.

\begin{figure}[t!]
\begin{minipage}[t]{0.99\textwidth}
  \begin{algorithm}[H]
    \caption{\textbf{Chunk-Causal Implementation.}}
    \small
    \begin{algorithmic}[1]
      \Require Chunk size $T$, input activation $X\in\mathbb R^{c_{\text{in}}\times nT\times h\times w}$, convolution kernel $W\in\mathbb R^{c_{\text{out}}\times c_{\text{in}}\times k_t\times k_h\times k_w}$
      \State Set $X \gets X.\text{reshape}(c_{\text{in}}, n, T, h, w).\text{transpose}(0, 1)\in \mathbb R^{n\times c_{\text{in}}\times T\times h\times w}$
      \State Set $\text{pad}_{t,\text{right}} \gets \mathbf{0} \in \mathbb R^{n\times c_{\text{in}}\times \frac{k_t-1}2\times h\times w}$
      \State Set $\text{pad}_{t,\text{left},\text{first}} \gets \mathbf{0} \in \mathbb R^{1\times c_{\text{in}}\times \frac{k_t-1}2\times h\times w}$
      \State Set $\text{pad}_{t,\text{left},\text{subsequent}} \gets x[:-1,:,-\frac{k_t-1}2:] \in \mathbb R^{(n-1)\times c_{\text{in}}\times \frac{k_t-1}2\times h\times w}$
      \State Set $\text{pad}_{t,\text{left}} \gets \text{concatenate}([\text{pad}_{t,\text{left},\text{first}}, \text{pad}_{t,\text{left},\text{subsequent}}],\text{dim}=0) \in \mathbb R^{n\times c_{\text{in}}\times \frac{k_t-1}2\times h\times w}$
      \State Set $X \gets \text{concatenate}([\text{pad}_{t,\text{left}},X,\text{pad}_{t,\text{right}}],\text{dim}=2) \in \mathbb R^{n\times c_{\text{in}}\times (T+k_t-1)\times h\times w}$
      \State Set $X \gets \text{spatial\_pad}(X, k_h-1, k_w-1) \in \mathbb R^{n\times c_{\text{in}}\times (T+k_t-1)\times (h+k_h-1)\times (w+k_w-1)}$
      \State Set $X \gets \text{convolution}(X, W) \in \mathbb R^{n\times c_{\text{out}}\times T\times h\times w}$
      \State Set $X \gets X.\text{transpose}(0, 1).\text{reshape}(c_{\text{out}}, nT, h, w) \in \mathbb R^{c_{\text{out}}\times nT\times h\times w}$
      \State \Return $X$
    \end{algorithmic}
    \label{alg:chunkcausal}
  \end{algorithm}
\end{minipage}
\end{figure}

Algorithm~\ref{alg:chunkcausal} presents the implementation details of our chunk-causal design. We mainly modify the temporal padding of the 3D convolution to enable chunk causal modeling. Specifically, for all chunks, the right temporal padding is set to zero to ensure causality between chunks, as shown in Algorithm~\ref{alg:chunkcausal} (line 2). For chunks other than the first chunk, their left temporal padding is set to the last \(\frac{k_t-1}{2}\) frames of the previous chunk to gather information, as shown in Algorithm~\ref{alg:chunkcausal} (line 4).

\myPara{Efficiency.}
\begin{table}[t]
\caption{
\textbf{Video Autoencoder Efficiency.} The encoding and decoding latency is measured for an $80 \times 256 \times 256$ video using an NVIDIA H100 GPU.
}
\vspace{-.05in}
\small\centering\setlength{\tabcolsep}{1pt}
\resizebox{0.9\linewidth}{!}{
\begin{tabular}{f | g | g | g g g | g g}
\toprule
\rowcolor{white} &  & Compress. & \multicolumn{3}{c|}{\#Params (M)} & \multicolumn{2}{c}{Latency (s)} \\
\rowcolor{white} \multirow{-2}{*}{Video Autoencoder} & \multirow{-2}{*}{Config} & Ratio & Total & Encoder & Decoder & Encode & Decode \\
\midrule
\rowcolor{white} VideoVAEPlus \cite{xing2024large}                    & f8t4c16   &  48 &  490.35 &  176.17 &  240.55 & 0.67 & 1.12 \\
\rowcolor{white} CogVideoX VAE \cite{yang2024cogvideox}               & f8t4c16   &  48 &  215.58 &   92.22 &  123.37 & 0.38 & 0.89 \\
\rowcolor{white} HunyuanVideo VAE \cite{kong2024hunyuanvideo}         & f8t4c16   &  48 &  246.48 &  100.34 &  146.14 & 0.46 & 0.81 \\
\rowcolor{white} IV VAE \cite{wu2025improved}                         & f8t4c16   &  48 &  241.85 &  101.70 &  140.15 & 0.55 & 0.84 \\
\rowcolor{white} Wan 2.1 VAE \cite{wan2025}                           & f8t4c16   &  48 &  126.89 &   53.59 &   73.29 & 0.27 & 0.43 \\
\rowcolor{white} Wan 2.2 VAE \cite{wan2025}                           & f16t4c48  &  64 &  704.69 &  149.63 &  555.05 & 0.15 & 0.43 \\
\rowcolor{white} StepVideo VAE \cite{ma2025step}                      & f16t8c64  &  96 &  498.82 &  109.89 &  388.93 & 0.23 & 0.69 \\
\midrule
\rowcolor{white} Video DC-AE$_{\scriptsize{\text{tiling \& blending}}}$ \cite{peng2025open} & f32t4c128 &  96 & 459.46 & 220.44 & 239.02 & 0.34 & 0.52 \\
\rowcolor{white} Video DC-AE \cite{peng2025open}                      & f32t4c128 &  96 &  459.46 &  220.44 &  239.02 & 0.45 & 0.69 \\
\rowcolor{white} LTX Video VAE \cite{hacohen2024ltx}                  & f32t8c128 & 192 & 1246.91 &  694.10 &  552.81 & 0.05 & 0.04 \\
\midrule
                                                                           & f32t4c256 &  48 & 2138.64 &  916.66 & 1221.98 & 0.34 & 0.36 \\
                                                                           & f32t4c128 &  96 & 2138.64 &  916.66 & 1221.98 & 0.34 & 0.36 \\
                                                                           & f32t4c64  & 192 & 2138.64 &  916.66 & 1221.98 & 0.34 & 0.36 \\
                                                                           & f32t4c32  & 384 & 2138.64 &  916.66 & 1221.98 & 0.34 & 0.36 \\
\multirow{-5}{*}{DC-AE-V}                                                  & f64t4c128 & 384 & 2717.28 & 1152.89 & 1564.39 & 0.34 & 0.36 \\
\bottomrule
\end{tabular}
}
\label{tab:video_ae_efficiency}
\end{table}

Table~\ref{tab:video_ae_efficiency} presents an efficiency comparison for various video autoencoders. The encoding and decoding latency is measured for an $80 \times 256 \times 256$ video using an NVIDIA H100 GPU. Our DC-AE-V achieves comparable latency to previous video autoencoders.

\clearpage

%
%
\bibliographystyle{splncs04}
\bibliography{main}
\end{document}